\documentclass[10pt,twocolumn,letterpaper]{article}

\usepackage{cvpr}              

\usepackage[dvipsnames]{xcolor}

\usepackage{epsfig}
\usepackage{graphicx}
\usepackage{amsmath}
\usepackage{amssymb}
\usepackage{booktabs}
\usepackage{bm}
\usepackage{multirow}
\usepackage[misc]{ifsym}
\usepackage[accsupp]{axessibility}

\definecolor{cvprblue}{rgb}{0.21,0.49,0.74}
\usepackage[pagebackref,breaklinks,colorlinks,citecolor=cvprblue]{hyperref}

\makeatletter
\def\UrlAlphabet{%
      \do\a\do\b\do\c\do\d\do\e\do\f\do\g\do\h\do\i\do\j%
      \do\k\do\l\do\m\do\n\do\o\do\p\do\q\do\r\do\s\do\t%
      \do\u\do\v\do\w\do\x\do\y\do\z\do\A\do\B\do\C\do\D%
      \do\E\do\F\do\G\do\H\do\I\do\J\do\K\do\L\do\M\do\N%
      \do\O\do\P\do\Q\do\R\do\S\do\T\do\U\do\V\do\W\do\X%
      \do\Y\do\Z}
\def\UrlDigits{\do\1\do\2\do\3\do\4\do\5\do\6\do\7\do\8\do\9\do\0}
\g@addto@macro{\UrlBreaks}{\UrlOrds}
\g@addto@macro{\UrlBreaks}{\UrlAlphabet}
\g@addto@macro{\UrlBreaks}{\UrlDigits}
\makeatother

\usepackage[capitalize]{cleveref}
\crefname{section}{Sec.}{Secs.}
\Crefname{section}{Section}{Sections}
\Crefname{table}{Table}{Tables}
\crefname{table}{Tab.}{Tabs.}

\newcommand\blfootnote[1]{%
  \begingroup
  \renewcommand\thefootnote{}\footnote{#1}%
  \addtocounter{footnote}{-1}%
  \endgroup
}

\def\paperID{****} 
\def\confName{CVPR}
\def\confYear{2024}

\title{SportsHHI: A Dataset for Human-Human Interaction Detection in Sports Videos}

\author{Tao Wu\textsuperscript{1,*} \quad \quad Runyu He\textsuperscript{1,*} \quad \quad  Gangshan Wu\textsuperscript{1} \quad \quad Limin Wang\textsuperscript{1,2,~\Letter}\\
$^1$State Key Laboratory for Novel Software Technology, Nanjing University \quad $^2$Shanghai AI Lab \\
{\tt\small  \{wt,runyu\_he\}@smail.nju.edu.cn, \{gswu, lmwang\}@nju.edu.cn} \\
\textbf{\normalsize\url{https://github.com/MCG-NJU/SportsHHI}}
}

\begin{document}
\maketitle
\begin{abstract}
Video-based visual relation detection tasks, such as video scene graph generation, play important roles in fine-grained video understanding. However, current video visual relation detection datasets have two main limitations that hinder the progress of research in this area. First, they do not explore complex human-human interactions in multi-person scenarios. Second, the relation types of existing datasets have relatively low-level semantics and can be often recognized by appearance or simple prior information, without the need for detailed spatio-temporal context reasoning. Nevertheless, comprehending high-level interactions between humans is crucial for understanding complex multi-person videos, such as sports and surveillance videos. To address this issue, we propose a new video visual relation detection task: video human-human interaction detection, and build a dataset named SportsHHI for it. SportsHHI contains 34 high-level interaction classes from basketball and volleyball sports. 118,075 human bounding boxes and 50,649 interaction instances are annotated on 11,398 keyframes. To benchmark this, we propose a two-stage baseline method and conduct extensive experiments to reveal the key factors for a successful human-human interaction detector. We hope that SportsHHI can stimulate research on human interaction understanding in videos and promote the development of spatio-temporal context modeling techniques in video visual relation detection.
\end{abstract}
\blfootnote{*: Equal contribution. ~~~\Letter: Corresponding author.}    
\section{Introduction}
\label{sec:intro}

Video understanding is a fundamental research field in computer vision, which has wide applications in security monitoring, internet video recommendation, and sports video analysis. Remarkable progress has been made in the action recognition task~\cite{ucf101,hmdb,i3d,kinetics400,kinetics600,kinetics700,something} with the advances of video convolution neural networks~\cite{s3d,c3d,i3d,r2p1d,slowfast,WangG18,tdn,TSN} and video transformers~\cite{vidtr,vivit,videoswin,mvit,VideoMAE,timesformer,videomaev2}. The action recognition task requires tagging each video clip with a single action label. However, real-world applications usually require a more detailed and in-depth understanding of videos, thus more attention has been attracted to fine-grained video understanding tasks such as video action detection~\cite{ava,avakinetics,multisports,jhmdb,acarn,aia,stmixer} and video visual relation detection tasks~\cite{AG,VidVRD,VidOR,sttran,jwj21}. 

\begin{figure}[t] 
    \centering
    \includegraphics[width=0.48\textwidth]{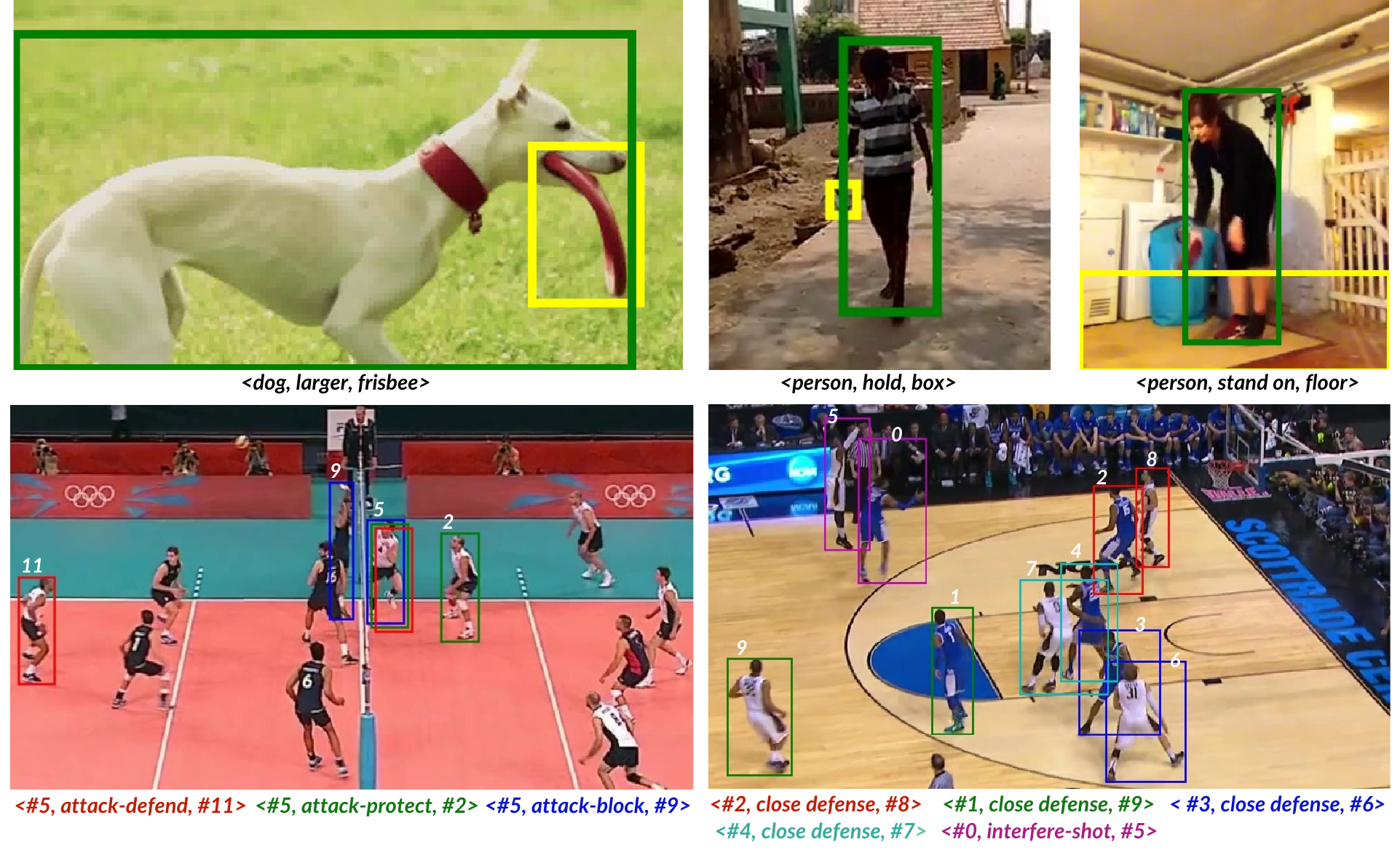}
    \vspace{-7mm}
    \caption{\textbf{Comparison between previous video visual relation detection datasets and our SportsHHI.} In the upper row, we show three relation instances from VidVRD and AG datasets. These datasets rarely involve human-human interaction and define semantically simple relations that can be recognized by appearance or prior information. In contrast, the bottom row shows interaction annotations in two sample keyframes of  SportsHHI. The bounding boxes and interaction annotation of the same instance are displayed in the same color. SportsHHI provides complex multi-person scenes where various interactions between human pairs occur concurrently. It focuses on high-level interactions that require detailed spatio-temporal context reasoning.} 
    \label{fig:samples}
    \vspace{-5mm}
\end{figure}

Video action detection involves localizing action performers in both space and time, as well as recognizing their action class. While earlier datasets like UCF101-24~\cite{ucf101} and JHMDB~\cite{jhmdb} primarily focuses on action detection in single-person scenarios, recent large-scale datasets such as AVA~\cite{ava} and MultiSports~\cite{multisports} have emphasized the importance of modeling human-human interactions for a better recognition of individual actions in multi-person scenes. However, while AVA and MultiSports provide ample multi-person scenes, neither of them offers explicit human-human interaction definitions and annotations.

Video scene graph generation is a popular video visual relation detection task, which requires the model to localize objects of interest in the video and recognize the relation of each pair of objects. While current datasets like AG~\cite{AG}, VidVRD~\cite{VidVRD}, and VidOR~\cite{VidOR} provide explicit relation instance annotations, human-human interactions are barely involved. Moreover, the relation categories defined in these benchmarks primarily include spatial relations, atomic actions, and simple visual comparisons, which are semantically straightforward and often recognizable by appearance information or simple prior knowledge, such as $\left \langle dog, larger, frisbee \right \rangle$ and $\left \langle person, stand\;on, floor \right \rangle$ in the upper row of Figure~\ref{fig:samples}. This makes current research pay more attention to the recognition of object categories and the utilization of category priors while ignoring reasoning based on spatio-temporal context. However, comprehending high-level interactions between humans is crucial for understanding complex multi-person videos, such as sports and surveillance videos. Spatio-temporal context modeling is an important technology that makes video relation detection different from image relation detection. We argue that there is a need for new datasets that comprehensively explore human-human interactions in complex multi-person scenarios, with high-level relation categories that require detailed spatio-temporal context reasoning.

In this paper, we propose a new video visual relation detection task: video human-human interaction detection. This task requires detecting human-human interaction instances in video frames. Each human-human interaction instance is formulated as a triplet $\left \langle S,I,O \right \rangle$, where $S$ and $O$ denote the bounding boxes of two different people and $I$ denotes the interaction category. For this task,  we develop a dataset named SportsHHI, short for Sports Human-Human Interaction. We label interaction instances on the keyframes of basketball and volleyball videos at 5FPS. Two samples of keyframes are shown in the bottom row of Figure~\ref{fig:samples}. SportsHHI has several unique characteristics that distinguish it from other visual relation detection datasets: 1) It is built on basketball and volleyball sports videos, thus containing a large number of complex multiplayer scenes with various interactions between athlete pairs occurring concurrently. 2) The interactions between each pair of humans are transient and change rapidly due to the fast movements of the players. 3) The defined interactions are of high-level semantic, including technical actions, tactical cooperation, and confrontation in sports. Recognizing an interaction instance in SportsHHI usually requires detailed spatio-temporal context reasoning. We hope that SportsHHI can attract more research attention to human-human interaction understanding in complex multi-person videos and promote the development of spatio-temporal context modeling techniques in video visual relation detection.

We test existing action detection and video scene graph generation methods on the SportsHHI dataset and propose a simple and neat baseline method based on these methods. Our baseline method adopts a Faster-RCNN-alike two-stage pipeline~\cite{fasterrcnn}: In the first stage, we use an offline human detector to detect person bounding boxes on the keyframe. We exhaust bounding box pairs to generate interaction proposals. In the second stage, we extract features for each proposal and categorize each proposal into an interaction class or background. Our experiments demonstrate that motion features, context information, relative position encoding, and information exchange among proposals are important for human-human interaction detection.

In summary, our contribution is three-fold: 1) We propose a new video relation detection task of human-human interaction detection which aims at exploring the complex interaction between people in multi-person scenarios; 2) We develop the SportsHHI dataset of multi-person sports videos on which high-level human-human interaction is well-defined and finely-labeled; 3) We design a two-stage baseline model and conduct extensive experiments on the SportsHHI dataset to discover the key factors for a successful interaction detector.

\section{Related Work}
\noindent \bf Video action detection. \rm
Video action detection aims to locate and recognize action instances on video frames. Earlier datasets like UCF Sports~\cite{ucf-sports}, UCF101-24~\cite{ucf101}, and J-HMDB~\cite{jhmdb} typically only include single-person videos. Datasets like DALY~\cite{daly}, AVA~\cite{ava}, AVA-Kinetics~\cite{avakinetics}, and MultiSports~\cite{multisports} contain scenes where multiple people are performing various actions concurrently. AVA sparsely annotates keyframes of movie videos at 1FPS. MultiSports collects videos from sports competitions. Many interaction-related action classes are defined, such as \textit{listen to} in AVA and \textit{second pass} in MultiSports. As they claim, it is important to model the interaction between people to recognize the action category of each person. However, as no interaction annotation is provided, interaction modeling can only be performed implicitly. SportsHHI annotates interactions on keyframes of basketball and volleyball videos from MultiSports at 5FPS. With interaction definitions and annotations provided in SportHHI, human-human interaction detection can be explicitly performed and evaluated.

\noindent \bf Video visual relation detection. \rm
Video scene graph generation requires localizing object pairs in the video and recognizing the relation of each pair of objects. AG~\cite{AG} is a large-scale dataset built on Charades~\cite{charades}. All videos in AG contain only one person, thus no human-human interaction is involved. In VidVRD~\cite{VidVRD} and VidOR~\cite{VidOR}, most of the videos contain only one person. Though human-human interaction is involved, the exploration of it is very limited. Similar to AG, video human-object interaction detection~\cite{st-hoi,vidor-hoid} datasets only focus on relations between humans and objects. Besides, the relation categories defined in these datasets only include spatial relations (\eg \textit{above}), atomic actions (\eg \textit{lean on}), and simple visual comparisons (\eg \textit{larger}). The relations are usually semantically simple and can be easily recognized by appearance information. Some trivial relations are also defined like $\left \langle dog, larger, frisbee \right \rangle$ in VidVRD. The category of the subject and object can often provide enough cues for relation recognition~\cite{trace,bsts,xfq2019,Tsai19,jwj21,sttran}, such as $\left \langle person, ride, horse \right \rangle$. Moreover, the relation between two objects changes very slowly in a video. 
Our SportHHI annotates human-human interaction instances in complex multi-person sports videos. The defined interactions are of high-level semantics and the interaction between two people changes rapidly.  

\noindent \bf Sports video understanding. \rm Researchers have built many different benchmarks in the sports domain for its challenges in spatio-temporal reasoning and promising application prospects, such as UCF101-24~\cite{ucf101} and MultiSports~\cite{multisports} for video action detection, FineGym~\cite{finegym} and Diving48~\cite{diving48} for action recognition and temporal action detection, SoccerNet~\cite{soccernet} for action spotting, SportsMOT~\cite{SportsMOT} for multi-object tracking, and NBA~\cite{NBA}, CAD~\cite{CAD} and Volleyball~\cite{volleyball} for group action recognition, etc. Group action recognition (GAR) requires tagging each video clip with a group action label. Implicit modeling of interactions among the players is often performed to improve the accuracy of label predictions. The proposed SportsHHI dataset provides interaction definitions and annotations for explicit human-human interaction exploration. 
\section{The SportsHHI Dataset}
In Sec.~\ref{sec:setcon}, we will explain several key choices we made while creating the SportsHHI dataset and outline the annotation process. Then in Sec.~\ref{sec:setsta}, we will present a detailed analysis of the statistics and characteristics of SportsHHI.

\subsection{Dataset Construction}
\label{sec:setcon}

\begin{figure}[t]
    \centering
    \includegraphics[width=0.47 \textwidth]{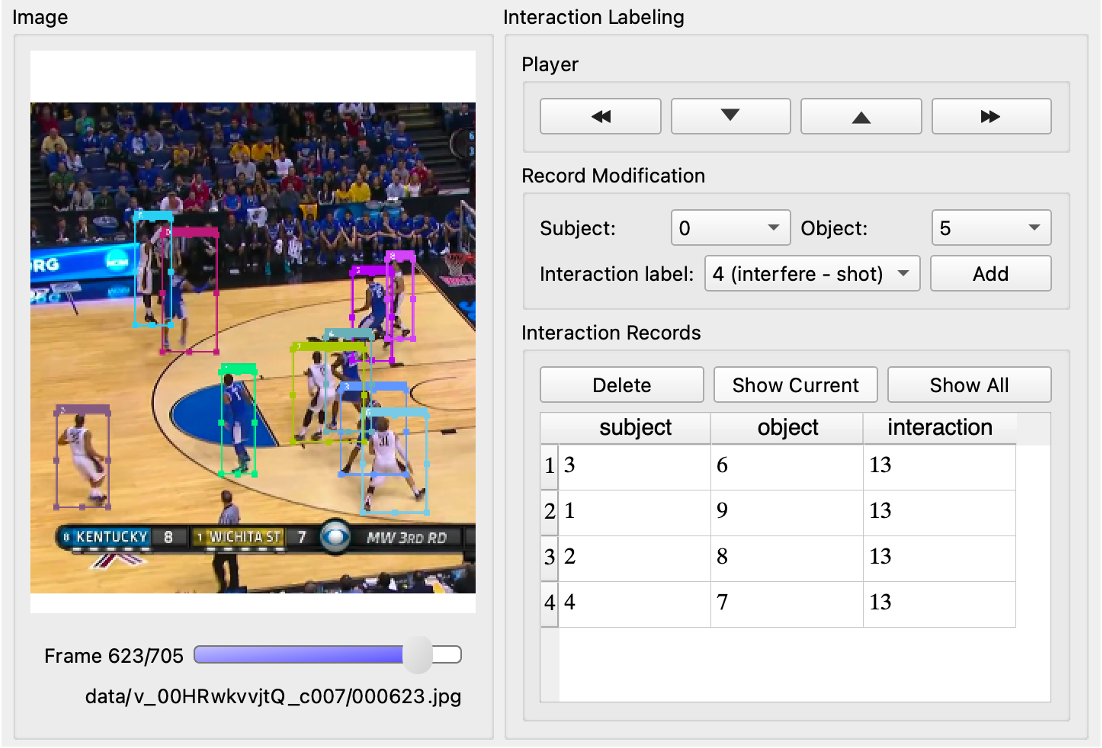}
    \vspace{-2mm}
    \caption{\textbf{User interface for interaction annotation.} The person bounding boxes and ids in the keyframe are shown in the left. We can play the video for context information. To add an interaction instance in the current keyframe, the subject person id, object person id, and interaction class should be specified.  }
    \label{fig:interface}
    \vspace{-2mm}
\end{figure}

\begin{figure}[t]
    \centering
    \includegraphics[width=0.48\textwidth]{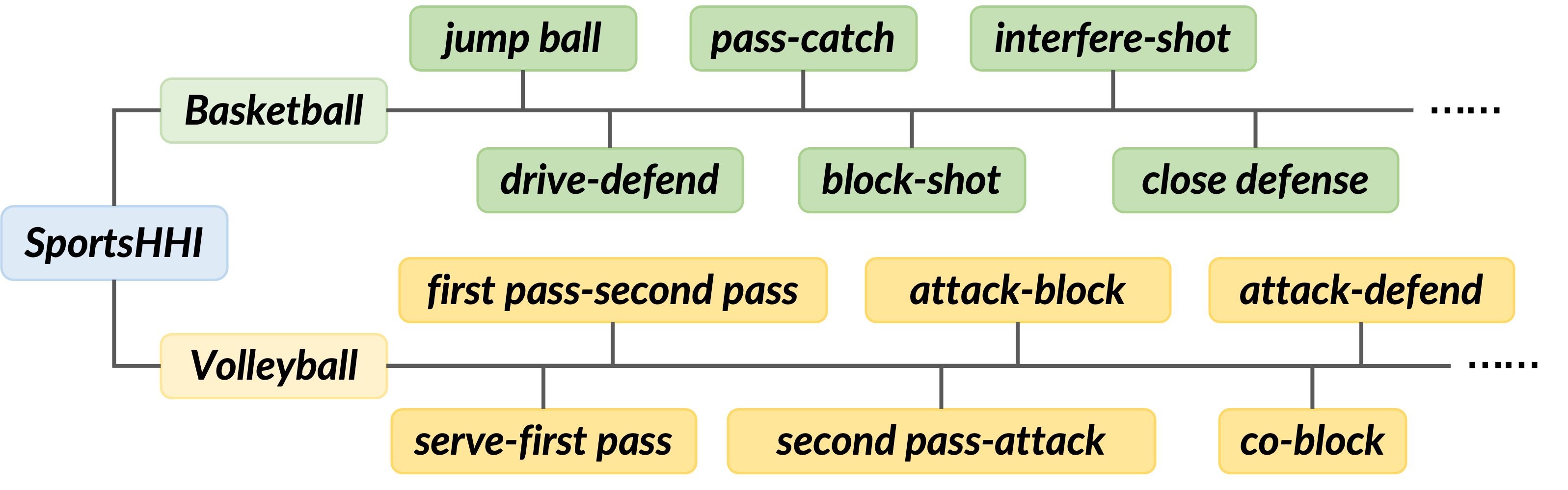}
    \vspace{-7mm}
    \caption{\textbf{Interaction classes hierarchy.} There are 34 interaction classes of high-level semantics in total in SportsHHI. 16 for basketball and 18 for volleyball. }
    \label{fig:classes}
    \vspace{-4mm}
\end{figure}

\begin{figure*}[htbp]
    \centering
    \includegraphics[width=16cm]{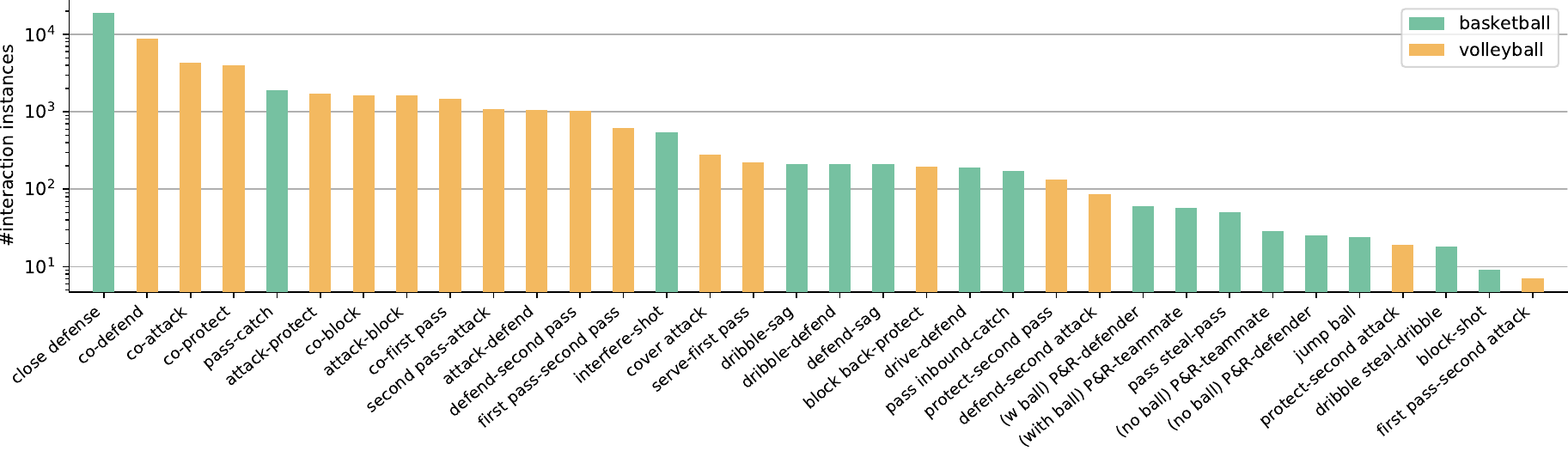}
    \vspace{-3mm}
    \caption{The number of interaction instances of each class sorted by descending order.  }
    \label{fig:insperclass}
    \vspace{-3mm}
\end{figure*}

\begin{figure}[htbp]
    \centering
    \includegraphics[width=8cm]{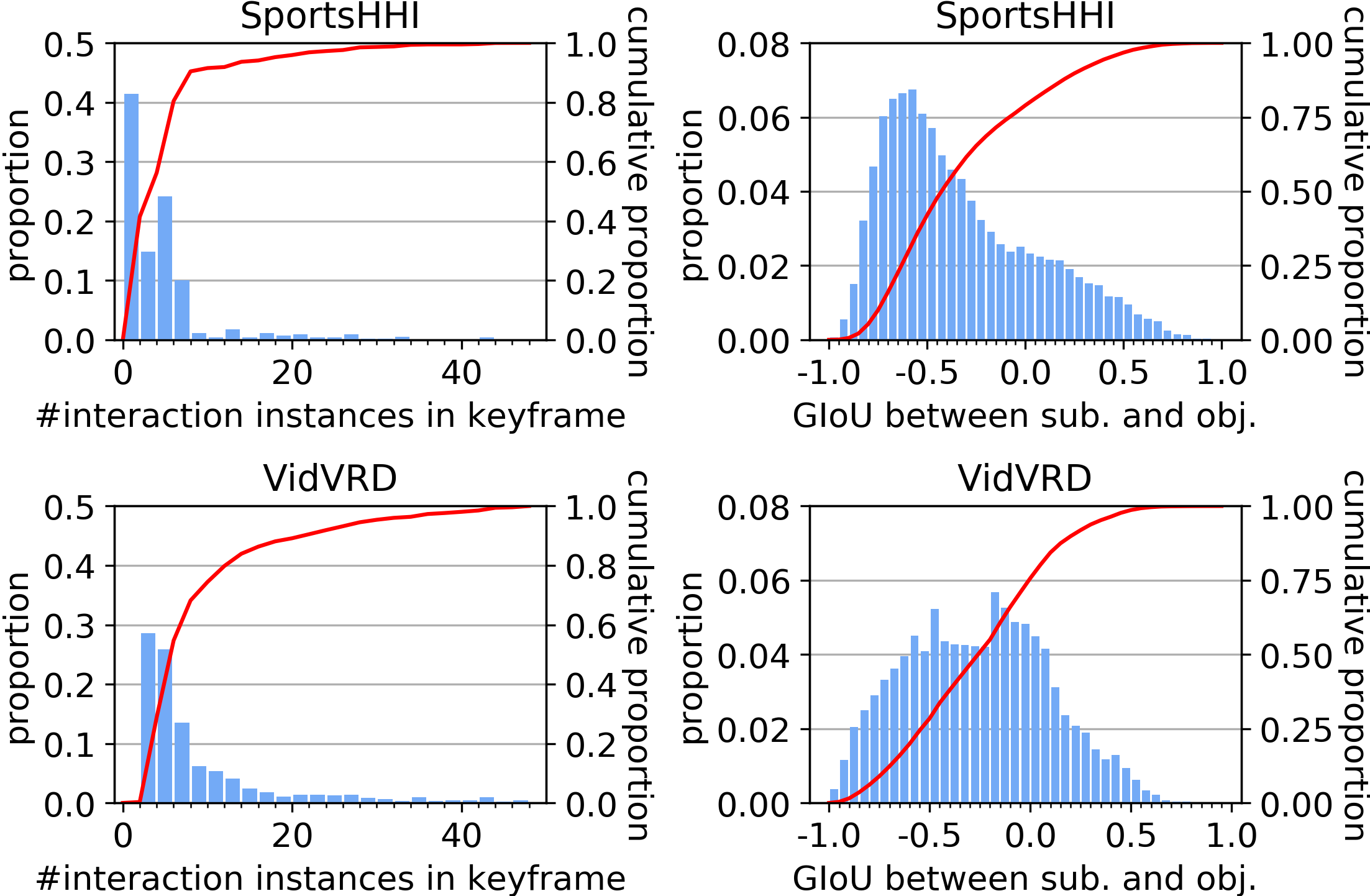}
    \vspace{-2mm}
    \caption{\textbf{Statistics comparisons between SportsHHI and VidVRD.} In the left, we compare the distribution of the number of instances in each keyframe. SportsHHI has more keyframes of fewer instances because of the high-level interaction class definition and the property of sports videos. In the right, we compare the distribution of GIoU between the subject and object. The proportion of instances of extremely high and extremely low GIoU between subject and object are both higher than VidVRD,  }
    \label{fig:insperframe}
    \vspace{-5mm}
\end{figure}
\noindent \bf Selection of the data domain. \rm 
First, we select to annotate human-human interaction in team sports videos for the following reasons:  1) The field of sports is itself a very promising application area for interaction detection. 2) Team sports videos provide ample multi-person scenes where various interactions between athlete pairs occur concurrently, which are rarely seen in daily-life videos. 3) Team sports videos involve many interactions with high-level semantics, such as tactics coordination or confrontation. 4) Interactions between people in sports game videos can be clearly and comprehensively defined according to game rules and professional athletes' advice. 5) High-quality, diverse sports game videos are easily accessible, while surveillance videos are usually of lower quality and harder to obtain because of privacy.  Furthermore, we decide to build the dataset on two popular team sports basketball and volleyball because: 1) The methods can be validated on two different subsets to evaluate their generalization performance on different sports. 2) Football is not included because the instance sparsity and class imbalance (which will be discussed in Sec.~\ref{sec:setsta}) would be more prominent in football videos.

\noindent \bf Interaction classes definition. \rm There are existing benchmarks focusing on atomic relations, which can be recognized by appearance information or prior knowledge. These interaction classes are out of our scope (it does not mean they are solved or not important). Instead, we focus on high-level interactions between athletes, which are semantically complex and require detailed spatio-temporal context reasoning to recognize. With the guidance of professional athletes, we generated the final interaction vocabulary through iterative trial labeling and modification, which is shown in Figure~\ref{fig:classes}. Our defined interaction classes include technical action (\eg \textit{pass-catch} in basketball), tactical coordination (\eg \textit{co-block} in volleyball), or confrontation (\eg \textit{attack-defend} in volleyball). 

\noindent \bf Interaction instance formulation. \rm Following common practice in AVA and AG datasets, we define interaction instances at the frame level, with reference to a long-term spatial-temporal context. Each interaction instance can be formulated as a triplet $\left \langle S,I,O \right \rangle$ where $S$ and $O$ denote the bounding boxes of the subject and object person and $I$ denotes the interaction category between them from the interaction vocabulary. When the subject person or the object person is out of view, we annotate $S$ or $O$ as ``invisible". This happens infrequently and we will provide statistics about it in the appendix.

\noindent \bf Data preparation. \rm
We carefully selected 80 basketball and 80 volleyball videos from the MultiSports~\cite{multisports} dataset to cover various types of games including men's, women's, national team, and club games. The average length of the videos is 603 frames and the frame rate of the videos is 25FPS. All videos have a high resolution of 720P.

\noindent \bf Annotation. \rm We follow the sparse annotation strategy of AVA~\cite{ava} and AG~\cite{AG} to reduce redundancy and save labor costs, but we annotate keyframes at a higher rate of 5FPS, because in sports videos, the change of interactions is very frequent and fast. In our practice, 5FPS can avoid lots of redundancy and keep up with the changes of interactions. 

We take a two-stage annotation pipeline. The first stage is for \textbf{person localization and tracking} and the second stage is for \textbf{interaction instance annotation}. The user interface of the software for interaction instance annotation is shown in Figure~\ref{fig:interface}. Though we annotate interaction instances at the frame level, as person id tracking is provided, we can easily generate interaction tubes by linking the same pair of persons with the same interaction class and provide temporal boundaries at a granularity of 5 frames. We will exemplify this in the appendix.
\begin{table}[]
\LARGE
\begin{center}
\resizebox{\linewidth}{!}{
\begin{tabular}{@{}l|c|c|c|c|c|c@{}}
\toprule
          & \#keyframes & \#interact. & \#inst. & \#obj. bbox & \#hum. bbox & avg. hum. \\ \midrule
AG~\cite{AG}        & 234253      & 25          & 1.7M    & 476229      & 234253      & 1.00      \\
VidVRD~\cite{VidVRD}    & 5834        & 134         & 55631   & 12705       & 2224        & 0.38      \\
SportsHHI & 11398       & 34          & 50649   &  -       &   118075          & 10.36          \\ \bottomrule
\end{tabular}
}
\end{center}
\vspace{-5mm}
\caption{Comparison of statistics between video scene graph generation datasets and SportsHHI. SportsHHI has a comparable scale to VidVRD. The average number of human bounding boxes per frame of SportsHHI is much higher than AG or VidVRD.}
\label{tab:statistics}
\vspace{-5mm}
\end{table}

\noindent \bf Quality control. \rm For the first stage, all annotations are complemented manually without using detection or tracking models. We double-check each video and manually correct inaccurate bounding boxes and inconsistent person numbers. For the second stage, all the annotators are professional athletes or veteran amateurs. To improve annotation accuracy and completeness, we perform repeated annotation. Each video is distributed to two different annotators. If the two annotators disagree on an annotation record, this record will be checked and decided by the meta-annotator.

\subsection{Dataset Statistics and Characteristics}
\label{sec:setsta}
Our SportsHHI provides interaction instance annotations on keyframes of basketball and volleyball videos. As shown in Table~\ref{tab:statistics}, SportHHI contains 34 interaction classes, and 11398 keyframes in total are annotated with instances. Current video scene graph datasets deal with general relations between various kinds of objects while our SportsHHI focuses on high-level interaction between humans. It is reasonable that datasets for video scene graph generation have a larger scale than SportsHHI. However, our SportHHI still has a comparable size to the popular VidVRD dataset for video scene graph generation. Our SportsHHI has more annotated keyframes (11398 versus 5834) and the number of interaction instances is close (55631 versus 50649). One important characteristic of our SportsHHI is the multi-person scenarios. The average number of people per frame in our SportsHHI is much higher than AG and VidVRD. AG only contains one person in each video and there is virtually no multi-person scenario in videos of VidVRD. Human-human interaction is barely involved in these datasets.

We further show some important characteristics of our SportsHHI: 1) As shown in Figure~\ref{fig:insperclass}, the distribution of the number of interaction instances per class roughly follows Zipf's law. This long-tail distribution requires us to put more attention to classes with fewer instances. 2) As shown in Figure~\ref{fig:insperframe} left, compared with VidVRD~\cite{VidVRD}, our SportsHHI has more keyframes with fewer instances. That is because we focus on interaction classes of coordination or confrontation in sports. In long plain segments (\eg segments of dribbling in basketball), few interactions of interest happen, and there are much more interaction instances in video segments with fierce confrontation. The videos are in crowd multi-person scenarios but the interaction instances are relatively sparse, which requires the model to distinguish two people without interaction from real interaction instances. 3) As shown in Figure~\ref{fig:insperframe} right, the proportion of instances of low GIoU between subject and object in SportsHHI is significantly higher than in VidVRD, which indicates that the subject and object of many instances are spatially far apart in SportsHHI. The proportion of extremely high GIoU is also higher, which indicates the occlusion is severe in some instances. VidVRD deals with simple relations in daily life videos, where the subject and object usually have a moderate distance.

\section{Method}\label{sec:method}

\begin{figure*}[t]
    \centering
    \includegraphics[width=\textwidth]{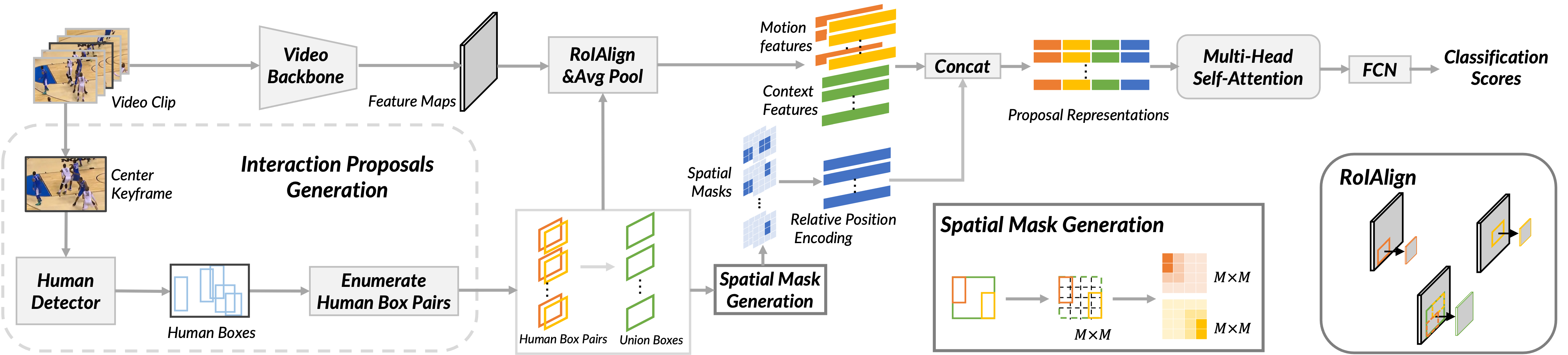}
    \caption{\textbf{An overview of our baseline method.} Given a video clip centered at a keyframe, we first use an offline human detector to detect human bounding boxes on the keyframe. All the human bounding box pairs are enumerated as interaction proposals. Then, we construct representations for the proposals. We perform the RoIAlign operation on the video feature maps extracted by a video backbone for motion features and context features. We generate spatial masks for each proposal, which are flattened and transformed into a vector. The vector is used as relative position encoding. The motion features, context features, and relative position encoding are concatenated as proposal representations.  Multi-head self-attention is applied to the representations for information exchange. Finally, an FCN classification head is used to predict the classification scores.}
    \label{fig:model}
    \vspace{-5mm}
\end{figure*}

We propose a baseline method for the human-human interaction detection task. As illustrated in Fig.~\ref{fig:model}, our method adopts a two-stage pipeline. In the first stage, interaction proposals are generated. In the second stage, we construct a representation for each proposal and classify each proposal into an interaction class or background. The following part of this section explains our method in detail.

\subsection{Interaction proposal generation}
We use an offline human detector for human detection on keyframes. We enumerate all human bounding box pairs as interaction proposals. Formally, for a keyframe at timestamp $t$, let $N$ denote the number of human boxes in the frame and $B =\left\{b_1, b_2, \cdots, b_N\right\}$ denote the boxes. The interaction proposals can be denoted as $P =\left\{p_1, p_2, \cdots, p_K\right\}$, where $p_k=(b_{s_k},b_{o_k})$ and $K=N\times (N-1)$.

\noindent \bf Positive proposals and negative proposals sampling. \rm We label a proposal $p_k$ as a positive proposal when its subject and object bounding boxes both have an IoU overlap higher than 0.7 with their counterparts in a ground-truth interaction instance. For training, it is possible to input all the interaction proposals to the classification network for loss function calculation, but this will consume large GPU memory and the network optimization will bias towards negative samples as they are dominant. Instead, we use all positive proposals and sample negative proposals by a fixed positive-negative ratio. To increase the proportion of positive proposals and avoid the situation that a ground truth has no corresponding positive proposal, we add all ground-truth human box pairs as positive proposals for training.

\subsection{Interaction classification}

The presentation of each interaction proposal comprises three parts: the motion features of the subject and object persons, the context features, and the relative position of the subject and object encoding. 

\noindent \bf Motion features and context features. \rm We sample a video clip centered at the keyframe and employ a 3D video backbone on it for feature map extraction. For an interaction proposal $p_k$, we apply RoIAlign operation~\cite{MaskRCNN} on the feature maps with the bounding boxes $b_{s_k}$ and $b_{o_k}$ and then transform the RoIAligned feature to dimension $d_v$ with a linear layer. The motion features are denoted as $f^{ms}_k\in \mathbb{R}^{d_v}$ and $f^{mo}_k\in \mathbb{R}^{d_v}$. We apply the RoIAlign operation on the features map using the union box of the subject and object for context information. The context feature is also transformed to dimension $d_v$, denoted as $f^c_k\in \mathbb{R}^{d_v}$

\noindent \bf Relative position information encoding. \rm Following~\cite{sttran,neuralmotifs}, given an interaction proposal $p_k$, we generate a spatial mask for the subject and object person respectively. We resize the union box of $b_{s_k}$ and $b_{o_k}$ to a fixed size of $M\times M$. We then generate the spatial mask of the subject bounding box $m_{k}^{s}\in \mathbb{R}^{M\times M}$. For each pixel in $m_{k}^{s}$, if it is inside the subject bounding box $b_{s_k}$, its value is set to 1, else 0. The spatial mask of the object bounding box $m_{k}^{o}\in \mathbb{R}^{M\times M}$ is generated in the same way. The two spatial masks are flattened, concatenated, and transformed to dimension $d_p$. We use the final vector $f^p_k \in \mathbb{R}^{d_p}$ as relative position encoding.

\noindent \bf Proposal representation. \rm We use the concatenation of motions features $f^{ms}_k$ and $f^{mo}_k$, context features $f^c_k$, and relative position encoding $f^p_k$ as the representation $x_k$ of interaction proposal $p_k$. Formally, 
\begin{equation}
  x_k = \left\langle f^{ms}_k, f^{mo}_k, f^c_k, f^p_k \right\rangle,  
\end{equation}
where $\left\langle,\right\rangle$ is the concatenation operation.

\noindent \bf Information exchange among proposals. \rm We denote the representation of the $K$ interaction proposals as $ X =\left\{\bm x_1, \bm x_2, \cdots, \bm x_K\right\}$. Sometimes, recognizing an interaction requires information from other interaction instances. For example, to recognize an interaction of \textit{co-defend} in volleyball, we need to know there exists an interaction of \textit{attack-defend}. To capture the information from other interaction proposals, we apply a standard multi-head self-attention (MHSA)~\cite{attention} operation on $X$. Formally,
\begin{equation}
    X^\prime = \text{MHSA}(X).
\end{equation}

\noindent \bf Classification head. \rm Finally, $X^\prime$ is input to a fully connected network (FCN) for classification score prediction. Let $S$ denote the predicted scores. Formally, 
\begin{equation}
     S = \text{FCN}(X^\prime),
\end{equation}
where $S\in \mathbb{R}^{K\times (C+1)}$, $C$ denotes the number of interaction classes. 

\section{Experiments}\label{sec:exp}
\subsection{Experimental Setup}
\noindent\bf Dataset. \rm We train and validate our model on the SportsHHI Dataset. The dataset is split into the training and validation set by video. Evaluation on a very small number of examples could be unreliable, so we evaluate the 28 classes that have at least 10 instances in both the training and validation set. Instances with an invisible subject or object are excluded from evaluation. In total, the current version contains 38,527 instances from 8,719 keyframes for training and 12,122 instances from 2,679 keyframes for validation. We report results on the validation set.

\noindent\bf Metrics. \rm A prediction is considered as a true positive if and only if its subject and object bounding boxes both have an IoU overlap higher than a preset threshold with their counterparts in a ground-truth interaction instance and the predicted interaction class matches the ground truth. Following VidVRD~\cite{VidVRD}, we set the IoU threshold to 0.5. Following the video scene graph generation task, we use Recall@K (K is the number of predictions) as the evaluation metric. Mean average precision is adopted as an evaluation metric for many detection tasks~\cite{coco}. However, this metric is discarded by many former visual relation detection benchmarks~\cite{AG,VG} because of their incomplete annotation. This issue does not exist in SportsHHI. In our experiments, mAP is also reported for the interaction detection task.  We argue that mAP is a more difficult and informative metric. Two different modes for model training and evaluation are used: 1)human-human interaction detection (HHIDet) which expects input video frames and predicts human boxes and interaction labels; 2) human-human interaction classification (HHICls) which directly uses gound-truth human bounding boxes and predicts human-human interaction classes.

\noindent\bf Implementation details. \rm The human detector we use is a Faster-RCNN~\cite{fasterrcnn} detector with a ResNeXt-101-FPN backbone, which is pre-trained on ImageNet~\cite{imagenet} and COCO~\cite{coco} and finetuned on keyframes of SportsHHI. The ratio of positive and negative proposals is set to 1:10. The SlowFast-R50 backbone~\cite{slowfast} pre-trained on Kinetics-400 ~\cite{kinetics400} is adopted for video feature extraction. The dimension of features $d_v$ and the dimension of relative position encoding $d_p$ are set to 512 and 256 respectively. We scale the short side of the video frames to 256. We use SGD optimizer with momentum 0.9 and weight decay $1 \times 10^{-5}$. The batch size is set to 8. The learning rate is 0.002 for HHICls and 0.004 for HHIDet. The models are trained for 20 epochs. 

\subsection{Ablation Study}
We conduct ablation experiments on SportsHHI to investigate the influence of each design and component of our baseline method. Unless otherwise stated, all ablation experiments are conducted in the HHICls mode, that is, the ground-truth human bounding boxes are used. HHICls mode allows us to study the key factors to predict interactions without the limitations of human detection. 

\noindent\bf Positive and negative proposal ratio. \rm We investigate the influence of the ratio of positive and negative proposals in Table~\ref{tab:gt_neg_pos}. A moderate proportion of negative samples help the model to distinguish interactions from backgrounds. However, when the positive-negative ratio gets too low, the model optimization will be overwhelmed by negative samples, and thus the performance drops.

\noindent\bf Spatial mask generation. \rm We generate two separate spatial masks for the subject and object person. Therefore, the positions of the subject and object can be distinguished. For comparison, we generate a single mask for the two people by setting the values in any box to 1. As shown in Table~\ref{tab:mask_abation}, model performance degrades when using a single mask. As the triplet $\left \langle S,I,O \right \rangle$ is ordered in SportsHHI, it is important to distinguish the position of the subject and object.

\noindent\bf Interaction proposal representation. \rm We show the importance of motion features $f^{mo}$ and $f^{ms}$, context feature $f^c$ and relative position encoding $f^p$ in interaction representation in Table~\ref{tab:gt_ablation}. We first show the results of using each feature in lines 1-3 and 8-10. Using only motions features $f^{mo}$ and $f^{ms}$ can achieve acceptable results, which indicates the modeling of the actions of the subject and object person is very important for the recognition of the interaction between them. The performance is much worse when using context feature $f^c$  or relative position encoding $f^p$ only. Using only the context feature cannot provide specific information of the subject and object and as the subject and object person are often far apart in spatial, the context feature may include much noise. Without visual information, pure prior information of the relative position of the subject and object person is not discriminative enough for interaction recognition. As shown in lines 5-7 and 12-14, when coupling position encoding with motion and context features, the performance improves a lot, which indicates that position information is complementary to visual information. As shown in lines 7 and 14, using all three types of features achieves the best results.

\noindent\bf Information exchange among proposals. \rm We ablate the multi-head self-attention module to show the importance of information exchange among proposals. As shown in Table~\ref{tab:gt_ablation}, regardless of the representation of the proposal, exchanging information among proposals always brings performance improvement.

\begin{table}[t]
\large
\begin{center}
\resizebox{0.95\linewidth}{!}{
\begin{tabular}{@{}ccc|c|ccccc@{}}
\toprule
$f^{mo}$\&$f^{ms}$ & $f^c$ & $f^p$ & Info. Ex. & mAP     & R@150 & R@100   & R@50 & R@20 \\ \midrule
\checkmark &-&-&-& 5.00 & 81.66 & 74.00 & 52.73 & 26.82 \\
-& \checkmark &-&-& 2.07 & 71.75 & 61.43 & 38.97 & 19.12 \\
-&-& \checkmark &-& 1.13 & 50.71 & 45.24 & 39.72 & 30.83 \\
\checkmark&\checkmark& - &-& 5.19 & 81.98 & 74.81 & 54.89 & 28.51 \\
\checkmark&-& \checkmark &-& 5.54 & 83.05 & 75.45 & 57.71 & 31.92 \\
-&\checkmark& \checkmark &-& 3.85 & 76.76 & 68.57 & 49.65 & 27.68 \\
\checkmark&\checkmark& \checkmark &-& 5.43 & 81.71 & 74.84 & 58.16 & \textbf{34.20}\\\hline
\checkmark &-&-&\checkmark& 6.15 & 84.09 & 76.59 & 55.55 & 28.10 \\
-& \checkmark &-&\checkmark& 3.51 & 73.41 & 63.11 & 41.47 & 20.91 \\
-&-& \checkmark &\checkmark& 2.01 & 58.76 & 51.60 & 38.88 & 26.35 \\
\checkmark&\checkmark& - &\checkmark& 6.82 & 84.01 & 77.11 & 58.14 & 31.86 \\
\checkmark&-& \checkmark &\checkmark& 6.94 & 84.41 & 76.72 & 57.44 & 31.05 \\
-&\checkmark& \checkmark &\checkmark& 5.97 & 80.75 & 71.82 & 51.78 & 31.04 \\
\checkmark&\checkmark& \checkmark &\checkmark& \textbf{7.52} & \textbf{85.78} & \textbf{78.52} & \textbf{59.53} & 32.76 \\\bottomrule
\end{tabular}}
\end{center}
\vspace{-6mm}
\caption{\textbf{Ablation study on the interaction representation and information exchanging.} ``\checkmark'' indicates the corresponding element is enabled while - indicates disabled. ``$f^{mo}$\&$f^{ms}$" denotes motion features of the subject and object person.``$f^c$" stands for context features of the union area. ``$f^p$" stands for the relative position information encoding. "Info. Ex." stands for the multi-head self-attention operation to exchange information among proposals.}
\label{tab:gt_ablation}
\vspace{-2.5mm}
\end{table}
\begin{table}[t]
\large
\begin{center}
\resizebox{0.7\linewidth}{!}{
\begin{tabular}{@{}c|ccccc@{}}
\toprule
Pos:Neg &  mAP   & R@150 & R@100 & R@50  & R@20  \\\midrule
1:1 & 6.97 & 85.92 & 79.64 & \textbf{60.96} & 32.59 \\
1:5 & 7.39 & \textbf{86.25} & 77.87 & 59.86 & \textbf{32.82} \\
1:10 & \textbf{7.52} & 85.78 & 78.52 & 59.53 & 32.76 \\
1:15 & 7.02 & 85.39 & 77.87 & 58.56 & 31.16\\
1:20 & 7.08 & 85.35 & 78.43 & 60.51 & 32.77 \\
1:25 & 6.89 & 85.44 & 78.31 & 58.85 & 31.61 \\
\bottomrule
\end{tabular}}
\end{center}
\vspace{-6mm}
\caption{\textbf{Ablation on the positive and negative proposals ratio.} The best results are obtained at 1:10. }
\label{tab:gt_neg_pos}
\vspace{-2.5mm}
\end{table}
\begin{table}[t]
\large
\begin{center}
\resizebox{0.75\linewidth}{!}{
\begin{tabular}{@{}l|ccccc@{}}
\toprule
 &  mAP   & R@150 & R@100 & R@50  & R@20  \\\midrule
Single Mask& 6.77 & 84.57 & 77.41 & 58.27 & 31.63 \\
Separate Mask& \textbf{7.52} & \textbf{85.78} & \textbf{78.52} & \textbf{59.53} & \textbf{32.76} \\\bottomrule
\end{tabular}}
\end{center}
\vspace{-6mm}
\caption{\textbf{Ablation on spatial mask generation.}``Separate Masks" means generating a spatial mask for the subject and object person respectively while "Single Mask" means using only one spatial mask to represent the pair.}
\label{tab:mask_abation}
\vspace{-2.5mm}
\end{table}
\begin{table}[t]
\large
\begin{center}
\resizebox{0.8\linewidth}{!}{
\begin{tabular}{@{}l|ccccc@{}}
\toprule
                      & mAP  & AP@0.5 & R@0.5 & R@0.7 & R@0.9 \\ \midrule
Human Boxes           & 73.5 & 92.6   & 98.6  & 96.0  & 58.2  \\
Interactions          & -    & -      & 98.4  & 93.2  & 37.7  \\ \bottomrule
\end{tabular}
}
\end{center}
\vspace{-6mm}
\caption{\textbf{Human detection and interaction proposal generation results.} For human box detection, we report mAP, AP, and Recall. For interaction proposal generation, Recall is reported. We calculate recall under IoU thresholds of 0.5, 0.7, and 0.9.}
\label{tab:stage1}
\vspace{-2mm}
\end{table}
\begin{table*}[t]
\begin{center}
\resizebox{.8\linewidth}{!}{
\begin{tabular}{l|l|ccccc|ccccc}
\toprule
\multirow{2}*{Method} & \multirow{2}*{Backbone} & \multicolumn{5}{c|}{HHICls} & \multicolumn{5}{c}{HHIDet}\\
\cline{3-12}
&& mAP & R@150 & R@100 & R@50 & R@20 & mAP & R@150 & R@100 & R@50 & R@20 \\\midrule
STTran~\cite{sttran} & ResNet-101~\cite{resnet} &  3.31 & 71.29 & 62.91 & 42.67 & 22.14 & 1.43 & 51.65 & 46.54 & 37.62 & 20.81  \\
HORT~\cite{jwj21} & ResNet-101~\cite{resnet} & 3.75 & 78.57 & 67.78 & 50.33 & 26.96
&  1.54 & 52.47 & 46.73 & 36.89 & 20/93  \\\hline
SlowFast~\cite{slowfast} & SlowFast-R50~\cite{slowfast} & 5.00 & 81.66 & 74.00 & 52.73 & 26.82 & 2.44 & 62.17 & 52.77 & 37.83 & 22.45 \\
ACARN~\cite{acarn} & SlowFast-R50~\cite{slowfast} & 5.44 & 82.85 & 75.22 & 56.53 & 31.77  & 2.53 & 62.25 & 52.84 & 37.50 & 21.85 \\ \hline
\multirow{3}{*}{Our baseline}  & SlowFast-R50~\cite{slowfast}  & 7.52 & 85.78 & 78.52 & 59.53 & 32.76 & 3.36 & 64.82 & 54.62 & 37.72 & 21.26\\
 & SlowFast-R101~\cite{slowfast} & 8.67 & 87.35 & 80.35 & 60.18 & 29.70 & 3.52 & 65.24 & 54.76 & 37.03 & 20.21 \\
 & ViT-B~\cite{VideoMAE} & \textbf{10.69} & \textbf{89.25} & \textbf{82.93} & \textbf{68.13} & \textbf{43.72} & \textbf{4.93} & \textbf{72.22} & \textbf{61.92} & \textbf{42.99} & \textbf{23.89} \\
\bottomrule
\end{tabular}
}
\end{center}
\vspace{-5.5mm}
\caption{\textbf{HHICls and HHIDet results.} We test existing video scene graph generation methods (STTran~\cite{sttran} and HORT~\cite{jwj21}) and action detection methods (SlowFast~\cite{slowfast} and ACARN~\cite{acarn}
) on SportsHHI. Our baseline method achieves the best performance. }
\label{tab:sota}
\vspace{-4mm}
\end{table*}

\subsection{Quantitative Results}

\noindent\bf Human box detection and interaction proposals generation results. \rm We first show the human detection and interaction results in Table~\ref{tab:stage1}. Under IoU threshold of 0.5, both human boxes and interactions have a high recall rate of 98.6 and 98.4. When the threshold improves to 0.7, recall of human boxes drops by 2.6 points and recall of interactions drops by 5.3 points. When improving the IoU threshold to 0.9, both recall rates drop significantly. The results indicate that, despite high recall under a relatively lower IoU threshold, there is still a lot of room for improvement in the quality of human boxes and interaction proposals. 

\noindent\bf HHICls and HHIDet results. \rm We provide HHICls and HHIDet results of some existing methods and our baseline in Table~\ref{tab:sota}. STTran~\cite{sttran} and HORT~\cite{jwj21} are two well-established video scene graph generation methods. They share two common properties: 1) Using appearance features extracted by image backbone. This is because the semantic level of relation classes defined by previous datasets is relatively low and the appearance feature is often sufficient for recognition, such as $\left \langle dog, larger, frisbee \right \rangle$. However, action modeling is very important for interaction recognition on SportsHHI. 2) Dependent on the object detection results. They add the category encoding of the objects to the relation representation, which provides strong prior information. For example, knowing that the subject and object are \textit{human} and \textit{horse} respectively, the relation category is very likely to be \textit{ride}. However, SportsHHI does not have such prior as all subjects and objects are humans. Since video backbones are used to extract the motion features of each person, the performance of the action detection methods like SlowFast~\cite{slowfast} and ACARN~\cite{acarn} is relatively better. Our baseline follows these action detectors to adopt a video backbone for motion features and introduce extra relative position and spatio-temporal context modeling between the subject and object person. Our simple baseline achieves leading performance with the same SlowFast-R50 backbone. When using the stronger VideoMAE~\cite{VideoMAE} pretrained ViT-B backbone, the performance is significantly improved, which shows the importance of spatio-temporal representation.

\subsection{Error Analysis}
\begin{figure}[t]
    \centering
    \includegraphics[width=0.46\textwidth]{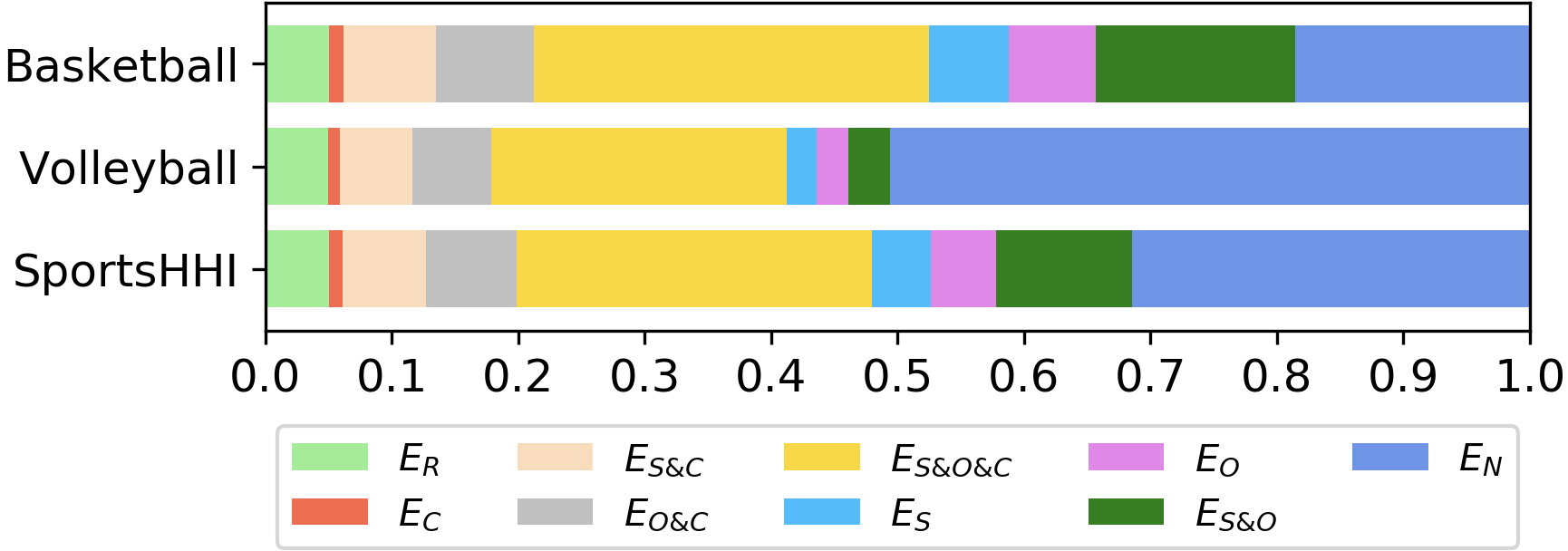}
    \vspace{-2mm}
    \caption{\textbf{Error analysis in HHIDet mode.} We show the proportion of each error type among false positives.}
    \label{fig:ea}
    \vspace{-4mm}
\end{figure}

Following ACT~\cite{ACT} and MultiSports~\cite{multisports}, we analyze error types of the false positives in the predictions to better understand the challenges of human-human interaction detection in SportsHHI. We analyze the predictions of the baseline model with a SlowFast-R50 backbone. We classify false positives in the predictions into 9 mutually exclusive error types. $E_R$: a prediction targets a ground-truth instance that has already been matched. $E_C$: a prediction that has an accurate subject and object localization, but wrong interaction class. $E_O$: a prediction that has accurate subject localization and correct interaction class, but inaccurate object localization. $E_S$: a prediction that has accurate object localization and correct interaction class, but inaccurate subject localization. $E_{S\&O}$, $E_{O\&C}$, $E_{S\&C}$, $E_{S\&O\&C}$: a prediction that is inaccurate in corresponding aspects while acceptable in other aspects (if any). $E_N$: a prediction that both the subject and object of it have no overlap with any ground truth. 

As shown in Figure~\ref{fig:ea}, $E_N$ is one of the most common error types. This is because SportsHHI is annotated in crowd multi-person scenarios but the interaction instances are relatively sparse due to the characteristic of sports videos and our high-level interaction definition. It is important but difficult to distinguish two people without interaction from real interaction instances. In video action detection~\cite{multisports}, localization is relatively accurate and most errors are about classification. However, in interaction detection, the localization of people involved in interaction and the classification of interaction are inseparable. Pure localization errors ($E_S$ and $E_O$) or pure classification errors ($E_C$) are relatively rare. The error types that coexisted with localization error and classification error ($E_{O\&C}$, $E_{S\&C}$, $E_{S\&O\&C}$) account for a relatively large proportion.  
\section{Conclusion}
In this paper, we proposed a new video visual relation detection task with a focus on sports human-human interaction understanding. This task deals with the complex interactions between humans in multi-person scenarios. We build a dataset named SportsHHI based on sports videos for this task. Human boxes and interaction instances are exhaustively annotated on keyframes at 5FPS. To benchmark this, we test several existing methods on SportsHHI and propose a baseline method. We conduct extensive experiments on SportsHHI and reveal the importance of motion information, context information, and position information for interaction recognition. We hope our SportHHI dataset and baseline method can inspire future research on human-human interaction understanding in videos.
\vspace{-2mm}
\paragraph {\bf Acknowledgements.} {\small This work is supported by National Key R$\&$D Program of China (No. 2022ZD0160900), National Natural Science Foundation of China (No. 62076119, No. 61921006)), and Collaborative Innovation Center of Novel Software Technology and Industrialization.}

\newpage
{
    \small
    \bibliographystyle{ieeenat_fullname}
    \bibliography{main}

\begin{thebibliography}{60}
\providecommand{\natexlab}[1]{#1}
\providecommand{\url}[1]{\texttt{#1}}
\expandafter\ifx\csname urlstyle\endcsname\relax
  \providecommand{\doi}[1]{doi: #1}\else
  \providecommand{\doi}{doi: \begingroup \urlstyle{rm}\Url}\fi

\bibitem[Arnab et~al.(2021)Arnab, Dehghani, Heigold, Sun, Lu{\v{c}}i{'c}, and Schmid]{vivit}
Anurag Arnab, Mostafa Dehghani, Georg Heigold, Chen Sun, Mario Lu{\v{c}}i{'c}, and Cordelia Schmid.
\newblock Vivit: A video vision transformer.
\newblock In \emph{ICCV}, pages 6836--6846, 2021.

\bibitem[Bertasius et~al.(2021)Bertasius, Wang, and Torresani]{timesformer}
Gedas Bertasius, Heng Wang, and Lorenzo Torresani.
\newblock Is space-time attention all you need for video understanding?
\newblock In \emph{{ICML}}, pages 813--824, 2021.

\bibitem[Carreira and Zisserman(2017)]{i3d}
Joao Carreira and Andrew Zisserman.
\newblock Quo vadis, action recognition? a new model and the kinetics dataset.
\newblock In \emph{CVPR}, pages 6299--6308, 2017.

\bibitem[Carreira et~al.(2018)Carreira, Noland, Banki-Horvath, Hillier, and Zisserman]{kinetics600}
Joao Carreira, Eric Noland, Andras Banki-Horvath, Chloe Hillier, and Andrew Zisserman.
\newblock A short note about kinetics-600.
\newblock \emph{arXiv preprint arXiv:1808.01340}, 2018.

\bibitem[Chiou et~al.(2021)Chiou, Liao, Wang, Zimmermann, and Feng]{st-hoi}
Meng{-}Jiun Chiou, Chun{-}Yu Liao, Li{-}Wei Wang, Roger Zimmermann, and Jiashi Feng.
\newblock {ST-HOI:} {A} spatial-temporal baseline for human-object interaction detection in videos.
\newblock In \emph{ICDAR@ICMR}, pages 9--17, 2021.

\bibitem[Choi et~al.(2009)Choi, Shahid, and Savarese]{CAD}
Wongun Choi, Khuram Shahid, and Silvio Savarese.
\newblock What are they doing? : Collective activity classification using spatio-temporal relationship among people.
\newblock In \emph{{ICCV} Workshops}, pages 1282--1289, 2009.

\bibitem[Cong et~al.(2021)Cong, Liao, Ackermann, Rosenhahn, and Yang]{sttran}
Yuren Cong, Wentong Liao, Hanno Ackermann, Bodo Rosenhahn, and Michael~Ying Yang.
\newblock Spatial-temporal transformer for dynamic scene graph generation.
\newblock In \emph{{ICCV}}, pages 16352--16362, 2021.

\bibitem[Cui et~al.(2023)Cui, Zeng, Zhao, Yang, Wu, and Wang]{SportsMOT}
Yutao Cui, Chenkai Zeng, Xiaoyu Zhao, Yichun Yang, Gangshan Wu, and Limin Wang.
\newblock Sportsmot: {A} large multi-object tracking dataset in multiple sports scenes.
\newblock In \emph{{ICCV}}, pages 9887--9897, 2023.

\bibitem[Deng et~al.(2009)Deng, Dong, Socher, Li, Li, and Fei-Fei]{imagenet}
Jia Deng, Wei Dong, Richard Socher, Li-Jia Li, Kai Li, and Li Fei-Fei.
\newblock Imagenet: A large-scale hierarchical image database.
\newblock In \emph{CVPR}, pages 248--255, 2009.

\bibitem[Fan et~al.(2021)Fan, Xiong, Mangalam, Li, Yan, Malik, and Feichtenhofer]{mvit}
Haoqi Fan, Bo Xiong, Karttikeya Mangalam, Yanghao Li, Zhicheng Yan, Jitendra Malik, and Christoph Feichtenhofer.
\newblock Multiscale vision transformers.
\newblock In \emph{ICCV}, pages 6824--6835, 2021.

\bibitem[Feichtenhofer et~al.(2019)Feichtenhofer, Fan, Malik, and He]{slowfast}
Christoph Feichtenhofer, Haoqi Fan, Jitendra Malik, and Kaiming He.
\newblock Slowfast networks for video recognition.
\newblock In \emph{ICCV}, pages 6202--6211, 2019.

\bibitem[Giancola et~al.(2018)Giancola, Amine, Dghaily, and Ghanem]{soccernet}
Silvio Giancola, Mohieddine Amine, Tarek Dghaily, and Bernard Ghanem.
\newblock Soccernet: {A} scalable dataset for action spotting in soccer videos.
\newblock In \emph{{CVPR} Workshops}, pages 1711--1721, 2018.

\bibitem[Goyal et~al.(2017)Goyal, Kahou, Michalski, Materzynska, Westphal, Kim, Haenel, Fr{\"{u}}nd, Yianilos, Mueller{-}Freitag, Hoppe, Thurau, Bax, and Memisevic]{something}
Raghav Goyal, Samira~Ebrahimi Kahou, Vincent Michalski, Joanna Materzynska, Susanne Westphal, Heuna Kim, Valentin Haenel, Ingo Fr{\"{u}}nd, Peter Yianilos, Moritz Mueller{-}Freitag, Florian Hoppe, Christian Thurau, Ingo Bax, and Roland Memisevic.
\newblock The ``something something" video database for learning and evaluating visual common sense.
\newblock In \emph{{ICCV}}, pages 5843--5851, 2017.

\bibitem[Gu et~al.(2018)Gu, Sun, Ross, Vondrick, Pantofaru, Li, Vijayanarasimhan, Toderici, Ricco, Sukthankar, et~al.]{ava}
Chunhui Gu, Chen Sun, David~A Ross, Carl Vondrick, Caroline Pantofaru, Yeqing Li, Sudheendra Vijayanarasimhan, George Toderici, Susanna Ricco, Rahul Sukthankar, et~al.
\newblock Ava: A video dataset of spatio-temporally localized atomic visual actions.
\newblock In \emph{CVPR}, pages 6047--6056, 2018.

\bibitem[He et~al.(2016)He, Zhang, Ren, and Sun]{resnet}
Kaiming He, Xiangyu Zhang, Shaoqing Ren, and Jian Sun.
\newblock Deep residual learning for image recognition.
\newblock In \emph{{CVPR}}, pages 770--778, 2016.

\bibitem[He et~al.(2017)He, Gkioxari, Doll{\'a}r, and Girshick]{MaskRCNN}
Kaiming He, Georgia Gkioxari, Piotr Doll{\'a}r, and Ross Girshick.
\newblock Mask r-cnn.
\newblock In \emph{ICCV}, pages 2961--2969, 2017.

\bibitem[Ibrahim et~al.(2016)Ibrahim, Muralidharan, Deng, Vahdat, and Mori]{volleyball}
Mostafa~S. Ibrahim, Srikanth Muralidharan, Zhiwei Deng, Arash Vahdat, and Greg Mori.
\newblock A hierarchical deep temporal model for group activity recognition.
\newblock In \emph{{CVPR}}, pages 1971--1980, 2016.

\bibitem[Jhuang et~al.(2013)Jhuang, Gall, Zuffi, Schmid, and Black]{jhmdb}
Hueihan Jhuang, Juergen Gall, Silvia Zuffi, Cordelia Schmid, and Michael~J Black.
\newblock Towards understanding action recognition.
\newblock In \emph{ICCV}, pages 3192--3199, 2013.

\bibitem[Ji et~al.(2020)Ji, Krishna, Fei-Fei, and Niebles]{AG}
Jingwei Ji, Ranjay Krishna, Li Fei-Fei, and Juan~Carlos Niebles.
\newblock Action genome: Actions as compositions of spatio-temporal scene graphs.
\newblock In \emph{CVPR}, pages 10236--10247, 2020.

\bibitem[Ji et~al.(2021)Ji, Desai, and Niebles]{jwj21}
Jingwei Ji, Rishi Desai, and Juan~Carlos Niebles.
\newblock Detecting human-object relationships in videos.
\newblock In \emph{{ICCV}}, pages 8086--8096, 2021.

\bibitem[Kalogeiton et~al.(2017)Kalogeiton, Weinzaepfel, Ferrari, and Schmid]{ACT}
Vicky Kalogeiton, Philippe Weinzaepfel, Vittorio Ferrari, and Cordelia Schmid.
\newblock Action tubelet detector for spatio-temporal action localization.
\newblock In \emph{{ICCV}}, pages 4415--4423, 2017.

\bibitem[Kay et~al.(2017)Kay, Carreira, Simonyan, Zhang, Hillier, Vijayanarasimhan, Viola, Green, Back, Natsev, et~al.]{kinetics400}
Will Kay, Joao Carreira, Karen Simonyan, Brian Zhang, Chloe Hillier, Sudheendra Vijayanarasimhan, Fabio Viola, Tim Green, Trevor Back, Paul Natsev, et~al.
\newblock The kinetics human action video dataset.
\newblock \emph{arXiv preprint arXiv:1705.06950}, 2017.

\bibitem[Kuehne et~al.(2011)Kuehne, Jhuang, Garrote, Poggio, and Serre]{hmdb}
Hildegard Kuehne, Hueihan Jhuang, Est{\'{\i}}baliz Garrote, Tomaso~A. Poggio, and Thomas Serre.
\newblock {HMDB:} {A} large video database for human motion recognition.
\newblock In \emph{{ICCV}}, pages 2556--2563, 2011.

\bibitem[Li et~al.(2020)Li, Thotakuri, Ross, Carreira, Vostrikov, and Zisserman]{avakinetics}
Ang Li, Meghana Thotakuri, David~A Ross, Jo{\~a}o Carreira, Alexander Vostrikov, and Andrew Zisserman.
\newblock The ava-kinetics localized human actions video dataset.
\newblock \emph{arXiv preprint arXiv:2005.00214}, 2020.

\bibitem[Li et~al.(2018)Li, Li, and Vasconcelos]{diving48}
Yingwei Li, Yi Li, and Nuno Vasconcelos.
\newblock {RESOUND:} towards action recognition without representation bias.
\newblock In \emph{{ECCV}}, pages 520--535, 2018.

\bibitem[Li et~al.(2021)Li, Chen, He, Wang, Wu, and Wang]{multisports}
Yixuan Li, Lei Chen, Runyu He, Zhenzhi Wang, Gangshan Wu, and Limin Wang.
\newblock Multisports: A multi-person video dataset of spatio-temporally localized sports actions.
\newblock In \emph{ICCV}, pages 13536--13545, 2021.

\bibitem[Lin et~al.(2014)Lin, Maire, Belongie, Hays, Perona, Ramanan, Doll{\'a}r, and Zitnick]{coco}
Tsung-Yi Lin, Michael Maire, Serge Belongie, James Hays, Pietro Perona, Deva Ramanan, Piotr Doll{\'a}r, and C~Lawrence Zitnick.
\newblock Microsoft coco: Common objects in context.
\newblock In \emph{ECCV}, pages 740--755. Springer, 2014.

\bibitem[Liu et~al.(2020)Liu, Jin, Xu, Gong, and Mu]{bsts}
Chenchen Liu, Yang Jin, Kehan Xu, Guoqiang Gong, and Yadong Mu.
\newblock Beyond short-term snippet: Video relation detection with spatio-temporal global context.
\newblock In \emph{{CVPR}}, pages 10837--10846, 2020.

\bibitem[Liu et~al.(2022)Liu, Ning, Cao, Wei, Zhang, Lin, and Hu]{videoswin}
Ze Liu, Jia Ning, Yue Cao, Yixuan Wei, Zheng Zhang, Stephen Lin, and Han Hu.
\newblock Video swin transformer.
\newblock In \emph{{CVPR}}, pages 3192--3201. {IEEE}, 2022.

\bibitem[Lu et~al.(2016)Lu, Krishna, Bernstein, and Fei{-}Fei]{VG}
Cewu Lu, Ranjay Krishna, Michael~S. Bernstein, and Li Fei{-}Fei.
\newblock Visual relationship detection with language priors.
\newblock In \emph{{ECCV}}, pages 852--869, 2016.

\bibitem[Pan et~al.(2021)Pan, Chen, Shou, Liu, Shao, and Li]{acarn}
Junting Pan, Siyu Chen, Mike~Zheng Shou, Yu Liu, Jing Shao, and Hongsheng Li.
\newblock Actor-context-actor relation network for spatio-temporal action localization.
\newblock In \emph{CVPR}, pages 464--474, 2021.

\bibitem[Qian et~al.(2019)Qian, Zhuang, Li, Xiao, Pu, and Xiao]{xfq2019}
Xufeng Qian, Yueting Zhuang, Yimeng Li, Shaoning Xiao, Shiliang Pu, and Jun Xiao.
\newblock Video relation detection with spatio-temporal graph.
\newblock In \emph{MM}, pages 84--93, 2019.

\bibitem[Ren et~al.(2015)Ren, He, Girshick, and Sun]{fasterrcnn}
Shaoqing Ren, Kaiming He, Ross Girshick, and Jian Sun.
\newblock Faster r-cnn: Towards real-time object detection with region proposal networks.
\newblock \emph{NIPS}, 28, 2015.

\bibitem[Rodriguez et~al.(2008)Rodriguez, Ahmed, and Shah]{ucf-sports}
Mikel~D. Rodriguez, Javed Ahmed, and Mubarak Shah.
\newblock Action {MACH} a spatio-temporal maximum average correlation height filter for action recognition.
\newblock In \emph{{CVPR}}, 2008.

\bibitem[Shang et~al.(2017)Shang, Ren, Guo, Zhang, and Chua]{VidVRD}
Xindi Shang, Tongwei Ren, Jingfan Guo, Hanwang Zhang, and Tat-Seng Chua.
\newblock Video visual relation detection.
\newblock In \emph{MM}, 2017.

\bibitem[Shang et~al.(2019)Shang, Di, Xiao, Cao, Yang, and Chua]{VidOR}
Xindi Shang, Donglin Di, Junbin Xiao, Yu Cao, Xun Yang, and Tat-Seng Chua.
\newblock Annotating objects and relations in user-generated videos.
\newblock In \emph{ICMR}, pages 279--287, 2019.

\bibitem[Shao et~al.(2020)Shao, Zhao, Dai, and Lin]{finegym}
Dian Shao, Yue Zhao, Bo Dai, and Dahua Lin.
\newblock Finegym: {A} hierarchical video dataset for fine-grained action understanding.
\newblock In \emph{{CVPR}}, pages 2613--2622, 2020.

\bibitem[Sigurdsson et~al.(2016)Sigurdsson, Varol, Wang, Farhadi, Laptev, and Gupta]{charades}
Gunnar~A. Sigurdsson, G{\"{u}}l Varol, Xiaolong Wang, Ali Farhadi, Ivan Laptev, and Abhinav Gupta.
\newblock Hollywood in homes: Crowdsourcing data collection for activity understanding.
\newblock In \emph{{ECCV}}, pages 510--526, 2016.

\bibitem[Smaira et~al.(2020)Smaira, Carreira, Noland, Clancy, Wu, and Zisserman]{kinetics700}
Lucas Smaira, Jo{\~a}o Carreira, Eric Noland, Ellen Clancy, Amy Wu, and Andrew Zisserman.
\newblock A short note on the kinetics-700-2020 human action dataset.
\newblock \emph{arXiv preprint arXiv:2010.10864}, 2020.

\bibitem[Soomro et~al.(2012)Soomro, Zamir, and Shah]{ucf101}
Khurram Soomro, Amir~Roshan Zamir, and Mubarak Shah.
\newblock {UCF101:} {A} dataset of 101 human actions classes from videos in the wild.
\newblock \emph{CoRR}, abs/1212.0402, 2012.

\bibitem[Sun et~al.(2021)Sun, He, Ren, and Wu]{vidor-hoid}
Xu Sun, Yunqing He, Tongwei Ren, and Gangshan Wu.
\newblock Spatial-temporal human-object interaction detection.
\newblock In \emph{{ICME}}, pages 1--6, 2021.

\bibitem[Tang et~al.(2020)Tang, Xia, Mu, Pang, and Lu]{aia}
Jiajun Tang, Jin Xia, Xinzhi Mu, Bo Pang, and Cewu Lu.
\newblock Asynchronous interaction aggregation for action detection.
\newblock In \emph{ECCV}, pages 71--87. Springer, 2020.

\bibitem[Teng et~al.(2021)Teng, Wang, Li, and Wu]{trace}
Yao Teng, Limin Wang, Zhifeng Li, and Gangshan Wu.
\newblock Target adaptive context aggregation for video scene graph generation.
\newblock In \emph{{ICCV}}, pages 13668--13677, 2021.

\bibitem[Tong et~al.(2022)Tong, Song, Wang, and Wang]{VideoMAE}
Zhan Tong, Yibing Song, Jue Wang, and Limin Wang.
\newblock Videomae: Masked autoencoders are data-efficient learners for self-supervised video pre-training.
\newblock In \emph{NeurIPS}, 2022.

\bibitem[Tran et~al.(2015)Tran, Bourdev, Fergus, Torresani, and Paluri]{c3d}
Du Tran, Lubomir Bourdev, Rob Fergus, Lorenzo Torresani, and Manohar Paluri.
\newblock Learning spatiotemporal features with 3d convolutional networks.
\newblock In \emph{ICCV}, pages 4489--4497, 2015.

\bibitem[Tran et~al.(2018)Tran, Wang, Torresani, Ray, LeCun, and Paluri]{r2p1d}
Du Tran, Heng Wang, Lorenzo Torresani, Jamie Ray, Yann LeCun, and Manohar Paluri.
\newblock A closer look at spatiotemporal convolutions for action recognition.
\newblock In \emph{{CVPR}}, pages 6450--6459, 2018.

\bibitem[Tsai et~al.(2019)Tsai, Divvala, Morency, Salakhutdinov, and Farhadi]{Tsai19}
Yao{-}Hung~Hubert Tsai, Santosh~Kumar Divvala, Louis{-}Philippe Morency, Ruslan Salakhutdinov, and Ali Farhadi.
\newblock Video relationship reasoning using gated spatio-temporal energy graph.
\newblock In \emph{{CVPR}}, pages 10424--10433, 2019.

\bibitem[Vaswani et~al.(2017)Vaswani, Shazeer, Parmar, Uszkoreit, Jones, Gomez, Kaiser, and Polosukhin]{attention}
Ashish Vaswani, Noam Shazeer, Niki Parmar, Jakob Uszkoreit, Llion Jones, Aidan~N. Gomez, Lukasz Kaiser, and Illia Polosukhin.
\newblock Attention is all you need.
\newblock In \emph{{NIPS}}, pages 5998--6008, 2017.

\bibitem[Wang et~al.(2016)Wang, Xiong, Wang, Qiao, Lin, Tang, and Gool]{TSN}
Limin Wang, Yuanjun Xiong, Zhe Wang, Yu Qiao, Dahua Lin, Xiaoou Tang, and Luc~Van Gool.
\newblock Temporal segment networks: Towards good practices for deep action recognition.
\newblock In \emph{ECCV}, pages 20--36, 2016.

\bibitem[Wang et~al.(2021)Wang, Tong, Ji, and Wu]{tdn}
Limin Wang, Zhan Tong, Bin Ji, and Gangshan Wu.
\newblock {TDN:} temporal difference networks for efficient action recognition.
\newblock In \emph{CVPR}, pages 1895--1904, 2021.

\bibitem[Wang et~al.(2023)Wang, Huang, Zhao, Tong, He, Wang, Wang, and Qiao]{videomaev2}
Limin Wang, Bingkun Huang, Zhiyu Zhao, Zhan Tong, Yinan He, Yi Wang, Yali Wang, and Yu Qiao.
\newblock Videomae {V2:} scaling video masked autoencoders with dual masking.
\newblock In \emph{CVPR}, pages 14549--14560, 2023.

\bibitem[Wang and Gupta(2018)]{WangG18}
Xiaolong Wang and Abhinav Gupta.
\newblock Videos as space-time region graphs.
\newblock In \emph{{ECCV}}, pages 413--431, 2018.

\bibitem[Weinzaepfel et~al.(2016)Weinzaepfel, Martin, and Schmid]{daly}
Philippe Weinzaepfel, Xavier Martin, and Cordelia Schmid.
\newblock Towards weakly-supervised action localization.
\newblock \emph{CoRR}, abs/1605.05197, 2016.

\bibitem[Wu et~al.(2019)Wu, Feichtenhofer, Fan, He, Krahenbuhl, and Girshick]{lfb}
Chao-Yuan Wu, Christoph Feichtenhofer, Haoqi Fan, Kaiming He, Philipp Krahenbuhl, and Ross Girshick.
\newblock Long-term feature banks for detailed video understanding.
\newblock In \emph{CVPR}, pages 284--293, 2019.

\bibitem[Wu et~al.(2020)Wu, Kuang, Wang, Zhang, and Wu]{carcnn}
Jianchao Wu, Zhanghui Kuang, Limin Wang, Wayne Zhang, and Gangshan Wu.
\newblock Context-aware rcnn: A baseline for action detection in videos.
\newblock In \emph{ECCV}, pages 440--456. Springer, 2020.

\bibitem[Wu et~al.(2023)Wu, Cao, Gao, Wu, and Wang]{stmixer}
Tao Wu, Mengqi Cao, Ziteng Gao, Gangshan Wu, and Limin Wang.
\newblock Stmixer: {A} one-stage sparse action detector.
\newblock In \emph{{CVPR}}, pages 14720--14729, 2023.

\bibitem[Xie et~al.(2018)Xie, Sun, Huang, Tu, and Murphy]{s3d}
Saining Xie, Chen Sun, Jonathan Huang, Zhuowen Tu, and Kevin Murphy.
\newblock Rethinking spatiotemporal feature learning: Speed-accuracy trade-offs in video classification.
\newblock In \emph{ECCV}, pages 305--321, 2018.

\bibitem[Yan et~al.(2020)Yan, Xie, Tang, Shu, and Tian]{NBA}
Rui Yan, Lingxi Xie, Jinhui Tang, Xiangbo Shu, and Qi Tian.
\newblock Social adaptive module for weakly-supervised group activity recognition.
\newblock In \emph{{ECCV}}, pages 208--224, 2020.

\bibitem[Zellers et~al.(2018)Zellers, Yatskar, Thomson, and Choi]{neuralmotifs}
Rowan Zellers, Mark Yatskar, Sam Thomson, and Yejin Choi.
\newblock Neural motifs: Scene graph parsing with global context.
\newblock In \emph{{CVPR}}, pages 5831--5840, 2018.

\bibitem[Zhang et~al.(2021)Zhang, Li, Liu, Shuai, Zhu, Brattoli, Chen, Marsic, and Tighe]{vidtr}
Yanyi Zhang, Xinyu Li, Chunhui Liu, Bing Shuai, Yi Zhu, Biagio Brattoli, Hao Chen, Ivan Marsic, and Joseph Tighe.
\newblock Vidtr: Video transformer without convolutions.
\newblock In \emph{ICCV}, pages 13577--13587, 2021.

\end{thebibliography}
}


\clearpage

\section*{Appendix}

\subsection*{A. More details about SportsHHI dataset}

\subsubsection*{A.1. Full interaction vocabulary}
We provide full interaction vocabulary of SportsHHI dataset in Table~\ref{tab:classes}. For comparison, we also provide the full relationship vocabulary of the popular AG dataset~\cite{AG} for video scene graph generation (VSGG) in Table~\ref{tab:AGclasses}. Our defined interaction classes are of high-level semantics, including technical action, tactical coordination, or confrontation while AG deals with low-level spatial relation or simple atomic action. In video scene graph generation~\cite{VidVRD,VidOR,VG}, the appearance features of the subject and object can often provide enough cues for relation inference~\cite{trace,bsts,xfq2019,Tsai19,jwj21,sttran}. For example, given that the subject is a person and the object is a clothes, it is highly possible the relation between them is ``wearing''. However, in human-human interaction detection, the subject and object are both person. Such prior information cannot be used for interaction recognition, and generally, it requires action modeling, relative position encoding and spatiotemporal context modeling.

\subsubsection*{A.2. Statistics of each sport}
We provide statistics of each sport in Table~\ref{tab:stateach}. The total number of instances of basketball and volleyball is close. Keyframe interaction instance distribution in basketball is more sparse than in volleyball because basketball videos have longer plain segments of dribbling. Both basketball and volleyball share the characteristics of crowd multi-person scenarios and relatively sparse interaction instance distribution, which requires the methods to distinguish two people without interaction from real interaction instances.

\subsubsection*{A.3. Statistics of partially invisible instances}
For an interaction instance $\left \langle S,I,O \right \rangle$, when the subject person or the object person is out of view, we annotate its $S$ or $O$ as ``invisible". This occurs due to two main reasons: camera angle switch and fast movement of the people. Table~\ref{tab:invisible} presents statistics of partially invisible interaction instances. In basketball, all partially invisible instances are from the interaction class of \textit{pass-catch}. This is because athletes move quickly during the transition between defense and offense, and sometimes the camera cannot fully capture the "pass-catch" process. In volleyball, partially invisible instances mainly occur in the interaction classes of \textit{serve-first pass} and \textit{co-first pass}. This is because sometimes the camera switches its focus from the serving player to capturing the entire court during the serve. The percentage of partially invisible instances is low because the high-level semantics we focus on are often also the focus of the view and can attract the camera's attention the most.

\begin{table}[!h]
\centering
\resizebox{\linewidth}{!}{
\begin{tabular}{@{}l|l@{}}
\toprule
\textbf{Basketball}                  & \textbf{Volleyball}        \\ \midrule
jump ball                            & serve - first pass         \\
pass - catch                         & co- first pass             \\
drive - defend                       & first pass - second pass   \\
block - shot                         & first pass - second attack \\
interfere - shot                     & co- attack                 \\
pass steal - pass                    & second pass - attack       \\
dribble steal - dribble              & cover attack               \\
dribble - defend                        & attack - block             \\
dribble - sag                     & co- block                  \\
defend - sag                         & attack - protect           \\
(with ball) pick-and-roll - defender & co- protect                \\
(with ball) pick-and-roll - teammate & block back - protect       \\
(no ball) pick-and-roll - defender   & protect - second pass      \\
(no ball) pick-and-roll - teammate   & protect - second attack    \\
pass inbound - catch                 & attack - defend            \\
close defense                        & co- defend                 \\
                                     & defend - second pass     \\
                                     & defend - second attack   \\
                                     \bottomrule
\end{tabular}
}
\caption{\textbf{Interaction vocabulary of SportsHHI}}
\label{tab:classes}
\end{table}
\begin{table}[!t]
\centering
\resizebox{\linewidth}{!}{
\begin{tabular}{@{}l|l|ll@{}}
\toprule
\textbf{attention} & \textbf{spatial} & \multicolumn{2}{c}{\textbf{contact}} \\ \midrule
looking at         & in front of      & carrying              & covered vy   \\
not looking at     & behind           & drinking from         & eating       \\
unsure             & on the side of   & leaning on            & holding      \\
                   & above            & have it on the back   & lying on     \\
                   & beneath          & not contacting        & sitting on   \\
                   & in               & standing on           & touching     \\
                   &                  & twisting              & wearing      \\
                   &                  & wiping                & writing on   \\ \bottomrule
\end{tabular}
}
\caption{\textbf{Relationship vocabulary of AG}}
\vspace{-1mm}
\label{tab:AGclasses}
\end{table}

\subsubsection*{A.4. Tube-level interaction instance generation}
\begin{figure*}[t]
  \centering
  \includegraphics[width=0.9\textwidth]{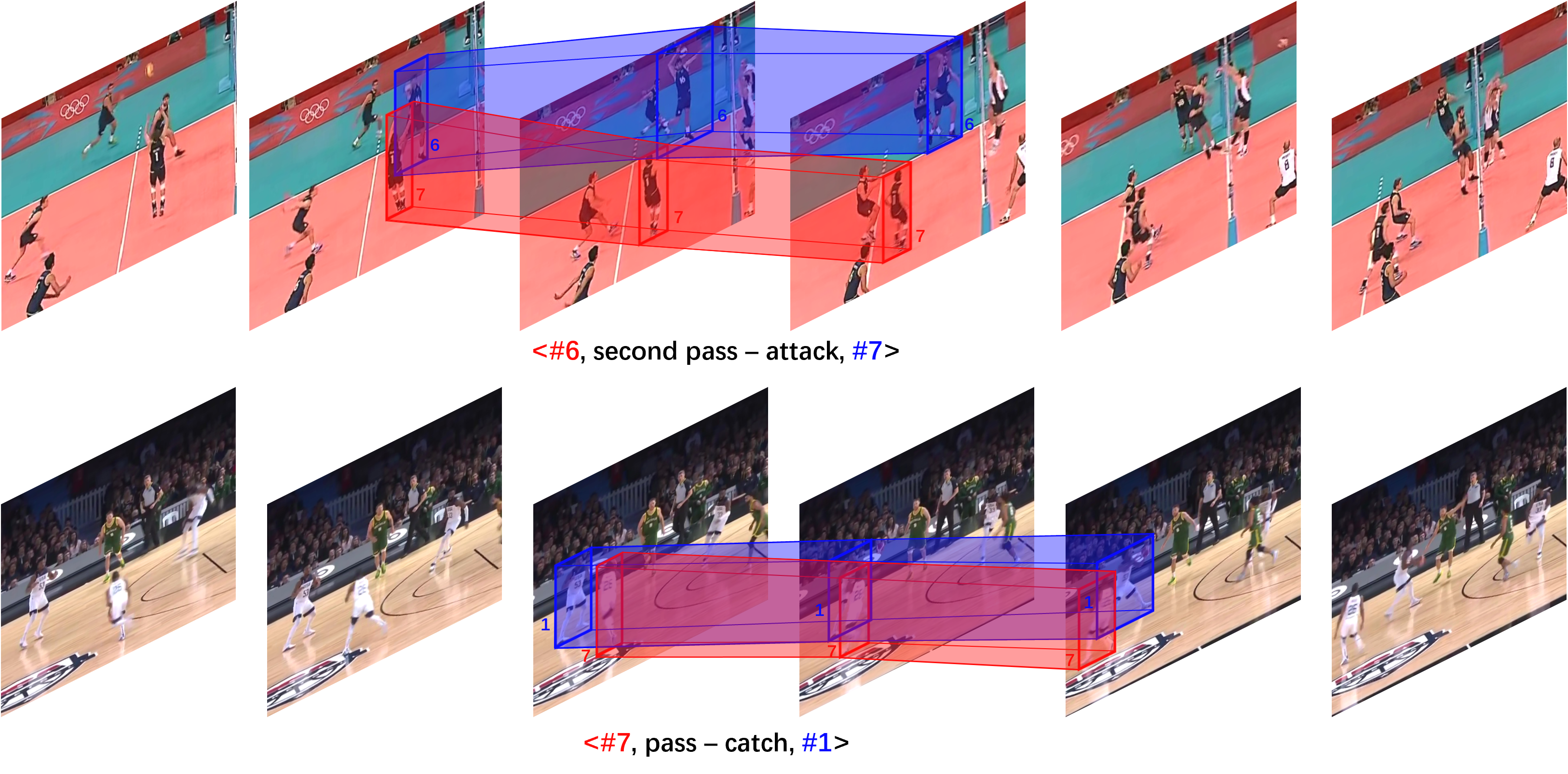}
  \vspace{-2mm}
  \caption{\textbf{Tube-level interaction instances.} We generate tube-level interaction instances by linking the same pair of people with the same interaction type across adjacent keyframes. The subjects are displayed in red and the objects in blue.  }
  \label{fig:tube}
\end{figure*}

We annotate interaction instances on the keyframes at 5FPS. Person id tracking is available, making it simple to create tube-level interaction instances by linking the same pair of individuals with the same interaction type across adjacent keyframes. Temporal boundaries can be provided at a granularity of 5 frames. Figure~\ref{fig:tube} illustrates two tube-level interaction instances generated using this approach. Most existing video visual relation detection benchmarks and methods focus on frame-level instance detection. We follow their lead and define interaction instances at the frame level. However, our SportsHHI can be readily adapted to tube-level video visual relation detection, which could potentially become a new research trend in the future.

\subsection*{B. More discussion about the baseline method}

\begin{table}[]
\centering
\resizebox{\linewidth}{!}{
\begin{tabular}{@{}l|c|c|c|c|c@{}}
\toprule
          & \#keyframes & \#interact. & \#inst.  & \#hum. bbox & avg. hum. \\ \midrule
Basketball        & 6791      & 16          & 22455       &  64549    & 9.51       \\
Volleyball    & 4607        & 18         & 28194        &  53526       & 11.62      \\ 
SportsHHI & 11398       & 34          & 50649        &   118075          & 10.36          \\ \bottomrule
\end{tabular}
}
\caption{\textbf{Statistics of SportsHHI}}
\label{tab:stateach}
\end{table}
\begin{table}[]
  \centering
  \begin{subtable}[t]{0.5\textwidth}
    \centering
    \resizebox{0.7\textwidth}{!}{
    \begin{tabular}{@{}l|c|c|c|c@{}}
    \toprule
    interact. class & \#inv. ins. & \#ins. & class \% & all \% \\ \midrule
    pass - catch    & 9                                     & 1894   & 0.48     & 0.04     \\
    others          & 0                                     & 0      & 0        & 0        \\ \bottomrule
    \end{tabular}}
    \vspace{-1mm}
    \caption{Basketball}
    \label{subtab:a}
  \end{subtable}
  \hfill
  \vspace{1mm}
  \begin{subtable}[t]{0.5\textwidth}
    \centering
    \resizebox{0.7\textwidth}{!}{
    \begin{tabular}{@{}l|c|c|c|c@{}}
    \toprule
    interact. class & \#inv. ins. & \#ins. & class \% & all \% \\ \midrule
    serve - first pass    & 13                                     & 225   & 5.78     & 0.05     \\
    co- first pass        & 52                                    & 1464      & 3.55        & 0.18        \\
    others          & 5                                     & -      & -        & 0.02        \\ \bottomrule
    \end{tabular}}
    \vspace{-1mm}
    \caption{Volleyball}
    \label{subtab:b}
  \end{subtable}
  \vspace{-2mm}
  \caption{\textbf{Statistics of partially invisible instances.} The table displays the count and percentage of partially invisible instances for each interaction class in the given sport. The percentage is calculated based on the total number of instances in that class as well as all instances in the sport.}
  \label{tab:invisible}
\end{table}

\begin{table*}[t]
\begin{center}
\resizebox{.8\linewidth}{!}{
\begin{tabular}{@{}ll|ccccc|ccccc@{}}
\toprule
\multirow{2}*{Training} & \multirow{2}*{Validation} & \multicolumn{5}{c|}{HHICls} & \multicolumn{5}{c}{HHIDet}\\
\cline{3-12}
&& mAP & R@150 & R@100 & R@50 & R@20 & mAP & R@150 & R@100 & R@50 & R@20 \\\midrule
Basketball & Basketball & 3.21 & 95.10 & 90.28 & 66.42 & 25.21 & 0.99 & 77.40 & 67.36 & 46.92 & 17.63\\
Volleyball & Volleyball & 15.65 & 83.52 & 74.85 & 58.11 & 34.06 & 8.09 & 65.18 & 54.04 & 36.21 & 20.35\\ \hline
SportsHHI  & Basketball & 3.52 & 94.94 & 91.44 & 78.16 & 52.97 & 1.39 & 79.80 & 71.35 & 53.35 & 29.22 \\
SportsHHI  & Volleyball & 15.82 & 84.23 & 75.41 & 59.05 & 34.00 & 7.65 & 65.52 & 53.61 & 34.29 & 18.25 \\
SportsHHI  & SportsHHI  & 10.69 & 89.25 & 82.93 & 68.13 & 43.72 & 4.93 & 72.22 & 61.92 & 42.99 & 23.89 \\
\bottomrule
\end{tabular}
}
\end{center}
\vspace{-2.5mm}
\caption{\textbf{HHICls and HHIDet results.} We show the results of training the model on the basketball or volleyball part of the SportsHHI training set and validating on the corresponding part of the validation set. We further show the results of training the model on the whole training set and validating on the basketball part, volleyball part, and whole validation set. ViT-B backbone is used.}
\label{tab:quant_res}
\vspace{-4mm}
\end{table*}

\subsubsection*{B.1. Comparisons with current Video VRD methods}
When designing the baseline method, we followed some practices of the current video scene graph generation and video human-object interaction detection methods, such as relative position encoding. However, there are still three major differences between current video visual relation detection (Video VRD) methods~\cite{trace,bsts,xfq2019,Tsai19,jwj21,sttran} and our baseline method: 1) They rely on appearance features extracted by image object detector and overlook motion modeling of the subject and object while we adopt 3D backbone for better modeling of each person's action. The semantic level of relation classes defined by previous datasets is relatively low and the appearance feature is often sufficient for recognition, such as $\left \langle dog, larger, frisbee \right \rangle$. However, action modeling is very important for interaction recognition on SportsHHI. For example, to distinguish between \textit{defend - second pass} and \textit{defend - second attack}, we need a good modeling of the object person's action to distinguish whether he is passing the ball or attacking. When we replace the motion features in interaction representation with appearance features, the performance drops significantly. 2) They are dependent on the accuracy of the image object detector in identifying object categories. For example, STTran~\cite{sttran} adds the category embedding of the subject and object to the relation representation, which provides strong prior information. For instance, knowing that the subject and object are \textit{human} and \textit{horse} respectively, the relation category is very likely to be \textit{ride}. However, SportsHHI does not have such a priori as all subjects and objects are humans. 3) Current Video VRD methods tend to treat different relation instances as independent individuals, while our baseline method exchanges information among interaction instances. In SportsHHI, sometimes, recognizing an interaction requires information from other interaction instances. For example, to recognize an interaction of \textit{co-defend} in volleyball, we need to know there exists an interaction of \textit{attack-defend}. 

\subsubsection*{B.2. Comparisons with action detection methods}

Current action detection methods~\cite{aia,acarn,lfb,carcnn} typically adopt the two-stage detection paradigm. Proposal person bounding boxes are generated in the first stage and RoI features are extracted for each proposal for action classification. Our baseline method for human-human interaction detection follows the two-stage pipeline. Unlike action detection methods, the RoI feature is insufficient for classification in the second stage. Experimental results show that context information, relative position encoding, and information exchange among the proposals are necessary for interaction recognition. Some methods model the interaction between people through attention mechanisms to improve the accuracy of action recognition of each individual person. However, the interaction modeling is implicitly performed as no interaction annotation is provided for supervision. The quality of interaction modeling can only be indirectly evaluated through the accuracy of action classification. In contrast, our baseline method deals with explicit human-human interaction modeling and prediction. The performance can be directly evaluated on the SportsHHI dataset.

\subsubsection*{B.3. Comparisons with action recognition methods}

Some methods for action recognition or group action recognition implicitly model the relations among people or objects to improve the action classification accuracy. For example, Wang et al.~\cite{WangG18} adopt a GCN to implicitly model relations among all people and objects in the video. Each node in their GCN is the appearance feature of a person or object and the nodes are connected according to similarity and adjacency. By performing graph convolution on the GCN, the information from all people and objects is gradually aggregated, which benefits the classification of the video. The goal of our baseline methods is different. Our method aims to identify whether there is an interaction between each pair of people and recognize the type of interaction. We adopt a Transformer~\cite{attention} to exchange information among interaction proposal representations to assist the recognition of each single proposal rather than aggregate global information for video-level classification.

\subsection*{C. More experimental results}

\begin{figure*}[htbp]
    \centering
    \includegraphics[width=.9\textwidth]{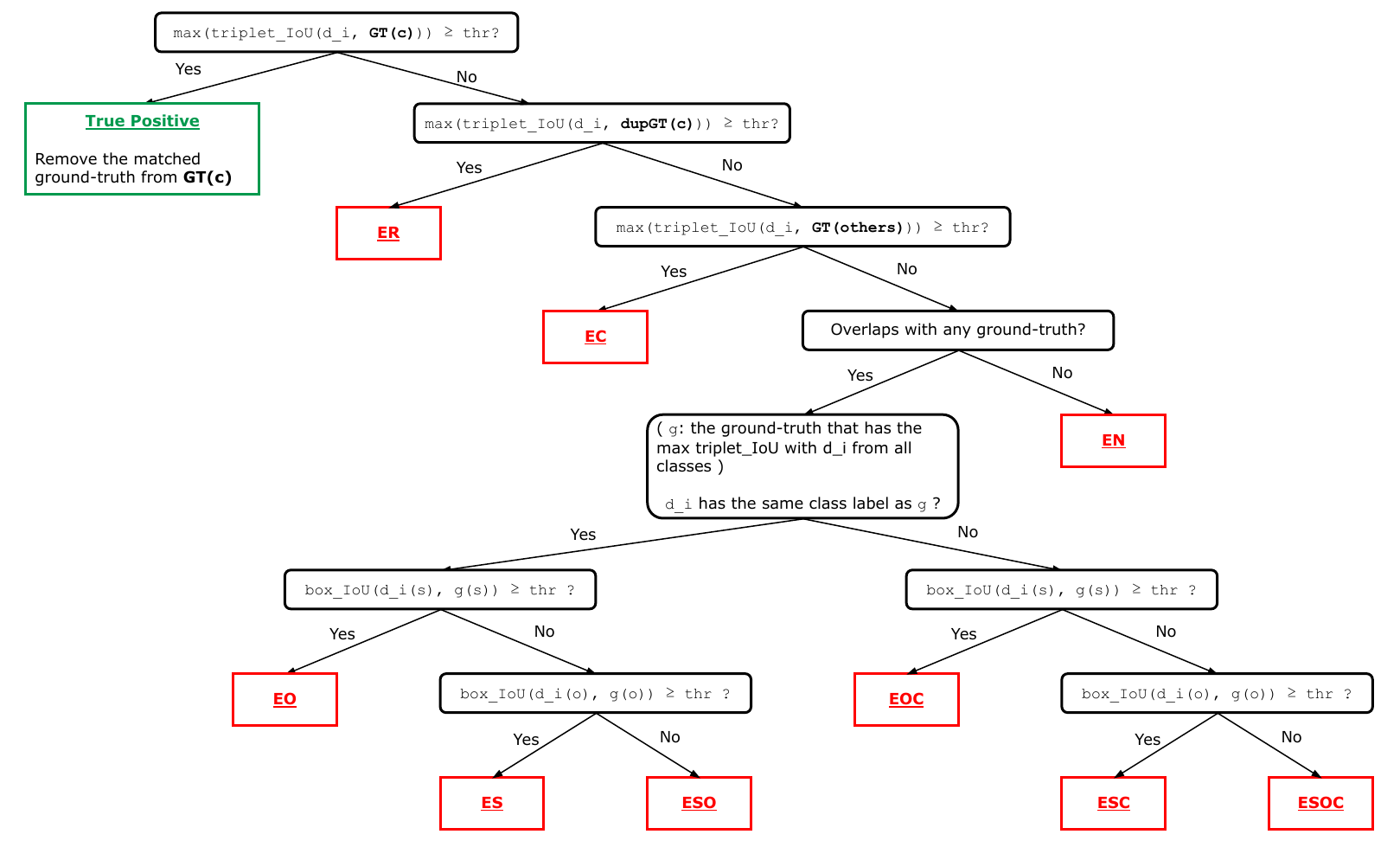}
    \vspace{-0.8em}
    \caption{\textbf{Error analysis tree.} For each detected triplet \texttt{d\_i} from a sorted list by descending order of confidence score of class \texttt{c}, \texttt{d\_i(s)} and \texttt{d\_i(o)} are the subject and object respectively. \texttt{g(s)} and \texttt{g(o)} are the subject and object for ground truth \texttt{g}. \texttt{box\_IoU} is the traditional IoU score for a pair of boxes. \texttt{triplet\_IoU} is the minimum of \texttt{box\_IoU} between subject boxes and object boxes. \texttt{\textbf{GT(c)}} is the set of ground truths of class \texttt{c}. \texttt{\textbf{dupGT(c)}} is the original copy of \texttt{\textbf{GT(c)}} and will not change during the error classification process.}
    \label{fig:ea_tree}
\end{figure*}

\noindent\bf HHICls and HHIDet results on each sport. \rm In Table~\ref{tab:quant_res}, we show the results of HHICls and HHIDet on each sport. We first show the result of training and validation on the basketball and volleyball parts of SportHHI respectively. Then we show the results of training on SportsHHI and validation on basketball, volleyball, and both. Using ground-truth human detection results, mAP and Recall of HHICls are higher than HHIDet with a large margin, which indicates the human detection results and the quality of interaction proposals have a significant influence on performance. Overall, our baseline method performs better on volleyball than basketball, because in basketball videos, the interaction patterns are more complex, and the interaction categories are more unbalanced. In the HHICls mode, compared with training the model on the subset of each sport, training on the whole training set of SportsHHI brings validation performance improvements on both basketball and volleyball. However, in the HHIDet mode, the improvement is not so stable. We speculate that, in the HHICls mode, due to the high quality of interaction proposals, through joint training, the model can learn more general representations. However, in the HHDet mode, low-quality proposals already introduce a lot of noise, and joint training further amplifies the impact of noise.

\subsection*{D. More detailed error analysis}

\begin{figure*}[!h]
    \centering
    \includegraphics[width=0.85\textwidth]{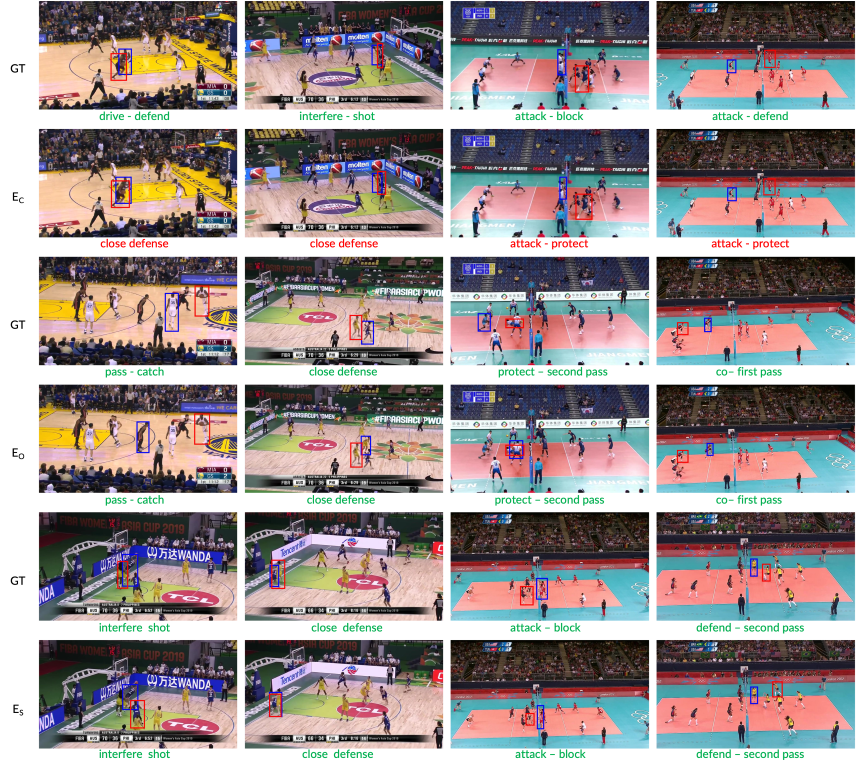}
    \vspace{-1.2em}
    \caption{\textbf{Visualization of typical errors for HHIDet on SportsHHI.} The subject person of a ground-truth or predicted interaction instance is marked in red and the object person is marked in blue. The ground-truth or correctly predicted interaction class labels are displayed in green and wrongly predicted in red. }
    \label{fig:eavis}
    \vspace{-1.8mm}
\end{figure*}

\begin{figure*}[!h]
    \centering
    \includegraphics[width=0.9\textwidth]{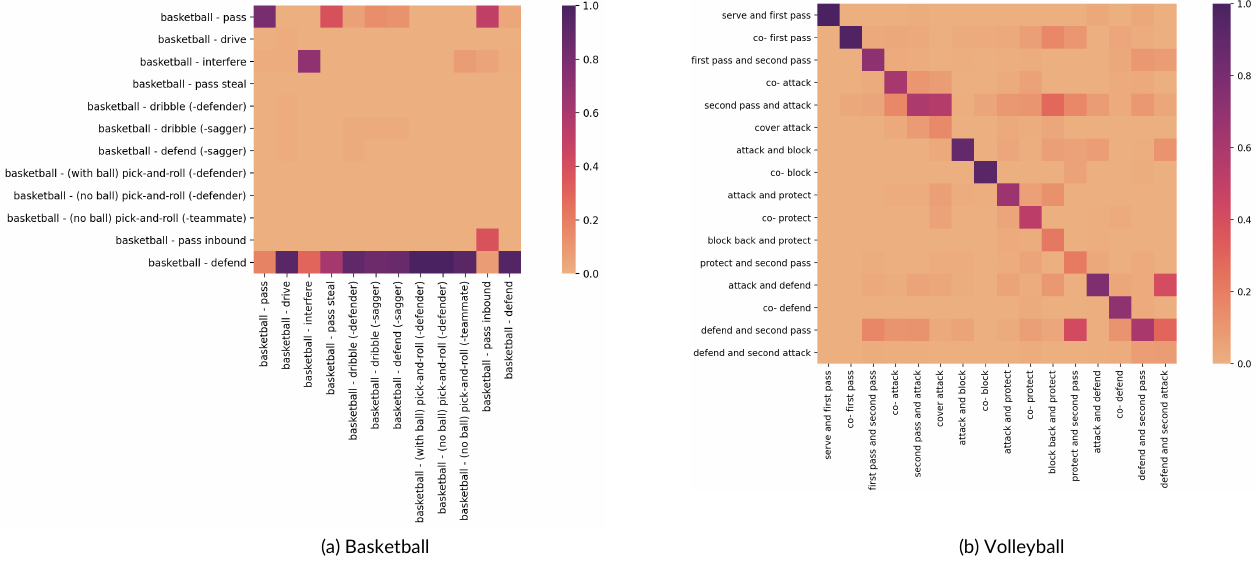}
    \vspace{-1.2em}
    \caption{\textbf{Confusion matrix of HHICls results on each sport}}
    \label{fig:confusion_matrix}
\end{figure*}

\subsubsection*{D.1 Error analysis tree}
Following ACT~\cite{ACT} and MultiSports~\cite{multisports}, we analyze error types of the false positives in the predictions to better understand the inherent difficulty in HHIDet. As illustrated in Figure \ref{fig:ea_tree}, we classify the detection errors into 9 mutually exclusive categories with a decision tree. A more detailed description of each error type is listed below.
\begin{itemize}
    \item $E_R$ (Errors of repeated detection): a detection result that has a triplet IoU larger than a threshold and the right action class with some ground truth triplet, but the ground truth triplet has already been matched by a detection result with a larger confidence score.
    \item $E_C$ (Errors of classification): a detection result that has the triplet IoU larger than a threshold with a ground truth, but its interaction class is not the same with the ground truth.
    \item $E_O$ (Errors of object localization): a detection result that has the same interaction class as a ground truth and the box IoU between the corresponding subject boxes are acceptable, but the box IoU between the object boxes are low.
    \item $E_S$(Errors of subject localization): a detection result that has the same interaction class as a ground truth and the box IoU between the corresponding object boxes are acceptable, but the box IoU between the subject boxes are low.
    \item $E_{S\&O}$ (Errors of subject and object localization): a detection result that has the same interaction class as a ground truth, but neither the object box IoU nor the subject box IoU meets the threshold.
    \item $E_{O\&C}$ (Errors of object localization and interaction classification): a detection result that has acceptable subject box IoU, but the object box IoU is low and the interaction class is incorrect.
    \item $E_{S\&C}$ (Errors of subject localization and interaction classification) a detection result that has acceptable object box IoU, but the subject box IoU is low and the interaction class is incorrect.
    \item $E_{S\&O\&C}$ (Errors of subject localization, object localization and interaction classification): a detection result that has low subject box IoU, low object box IoU and the interaction class is incorrect.
    \item $E_N$ (Errors of not matched): a detection result that has no overlap with any ground truth triplets of any class, indicating there should be no interaction detection results.
\end{itemize}

\subsubsection*{D.2. Visualization of error analysis}
We provide visualizations of false positives of some error types in Figure~\ref{fig:eavis}
to show the challenge of human-human interaction detection on SportsHHI more intuitively.

\subsubsection*{D.3. Confusion Matrix}
We draw the confusion matrix of HHICls predictions in Figure \ref{fig:confusion_matrix}. We observe that the model generally performs better on volleyball classes than on basketball classes. This is because the interaction patterns are more complicated and the number of instances of each class is more unevenly distributed in basketball. From Figure~\ref{fig:confusion_matrix}, we observe the challenges of SportsHHI in the following three aspects.

\begin{enumerate}
    \item \textbf{Handling long-tail distribution.} In basketball data for example, classes like \textit{sag} or \textit{pick-and-roll} have only a small number of instances while \textit{close defense} is very common. The optimization of the interaction classification network will be biased towards the dominant classes. However, the long-tail distribution is natural and inevitable in real-world data and how to catch the rare interactions remains a difficult yet important problem.
    \item \textbf{Action modeling.} Confusion between volleyball classes \textit{defend - second pass} and \textit{defend - second attack} indicates the importance of the action modeling of each person in SportsHHI. To distinguish between these two classes, we need to accurately identify whether the object person's action is ``pass" or ``attack". Former video visual relation detection datasets do not emphasize modeling human actions, so the current methods only used appearance features, which is not sufficient for interaction recognition on SportsHHI.
    \item \textbf{Long-term temporal structure modeling.} As a baseline model, we only leverage spatiotemporal context with a short video clip for neatness and simplicity. However, in order to distinguish between classes like \textit{protect - second pass} and \textit{defend - second pass}, we need longer temporal information to distinguish whether the ball is defended or protected. 
\end{enumerate}

\end{document}